\documentclass{article}

\usepackage[preprint]{neurips_2023}

\usepackage[utf8]{inputenc} 
\usepackage[T1]{fontenc}    
\usepackage{hyperref}       
\usepackage{url}            
\usepackage{booktabs}       
\usepackage{amsfonts}       
\usepackage{nicefrac}       
\usepackage{microtype}      
\usepackage{xcolor}         
\usepackage{graphicx}
\usepackage{amsmath}

\DeclareMathOperator*{\argmax}{argmax}

\newtheorem{theorem}{Theorem}

\newtheorem{lemma}[theorem]{Lemma}
\newtheorem{definition}{Definition}
\newtheorem{remark}{Remark}

\usepackage{times}
\usepackage{microtype}
\usepackage{epsfig}
\usepackage{xcolor}
\usepackage{caption}
\usepackage{float}
\usepackage{placeins}
\usepackage{color, colortbl}
\usepackage{stfloats}
\usepackage{enumitem}
\usepackage{tabularx}
\usepackage{xstring}
\usepackage{multirow}
\usepackage{xspace}
\usepackage{url}
\usepackage{subcaption}
\usepackage{xcolor}
\usepackage[hang,flushmargin]{footmisc}

% ...

\newcommand{\cN}{\mathcal{N}\xspace}

\newcommand{\cP}{\mathcal{P}\xspace}

\title{Scalar Invariant Networks with Zero Bias}

\author{%
  David S.~Hippocampus\thanks{Use footnote for providing further information
    about author (webpage, alternative address)---\emph{not} for acknowledging
    funding agencies.} \\
  Department of Computer Science\\
  Cranberry-Lemon University\\
  Pittsburgh, PA 15213 \\
  \texttt{hippo@cs.cranberry-lemon.edu} \\
}

\author{%
  Chuqin Geng \\
  Department of Computer Science\\
  McGill University\\
  Montreal, QC, Canada H3A 0G4 \\
  \texttt{chuqin.geng@mail.mcgill.ca} \\
  \And
  Xiaojie Xu \\
  Department of Computer Science \\
  McGill University \\
  Montreal, QC, Canada H3A 0G4 \\
  \texttt{xiaojie.xu@mail.mcgill.ca} \\
  \AND
  Haolin Ye \\
  Department of Computer Science \\
  McGill University  \\
  Montreal, QC, Canada H3A 0G4 \\
  \texttt{haolin.ye@mail.mcgill.ca} \\
  \And
  Xujie Si \\
  Department of Computer Science \\
  University of Toronto \\
  Toronto, ON, Canada M5S 1A1 \\
  \texttt{six@cs.toronto.edu} \\
}

\begin{document}

\maketitle

\begin{abstract}
Just like weights, bias terms are the learnable parameters of many popular machine learning models, including neural networks. Biases are thought to enhance the representational power of neural networks, enabling them to solve a variety of tasks in computer vision. However, we argue that biases can be disregarded for some image-related tasks such as image classification, by considering the intrinsic distribution of images in the input space and desired model properties from first principles. Our findings suggest that zero-bias neural networks can perform comparably to biased networks for practical image classification tasks. We demonstrate that zero-bias neural networks possess a valuable property called scalar (multiplication) invariance. This means that the prediction of the network remains unchanged when the contrast of the input image is altered. We extend scalar invariance to more general cases, enabling formal verification of certain convex regions of the input space. Additionally, we prove that zero-bias neural networks are fair in predicting the zero image. Unlike state-of-the-art models that may exhibit bias toward certain labels, zero-bias networks have uniform belief in all labels. We believe dropping bias terms can be considered as a geometric prior in designing neural network architecture for image classification, which shares the spirit of adapting convolutions as the transnational invariance prior. The robustness and fairness advantages of zero-bias neural networks may also indicate a promising path towards trustworthy and ethical AI.

\end{abstract}

\section{Introduction}
\label{sec:intro}

\begin{figure}
     \centering
     \begin{subfigure}[b]{0.31\textwidth}
         \centering
            \includegraphics[width=\textwidth]{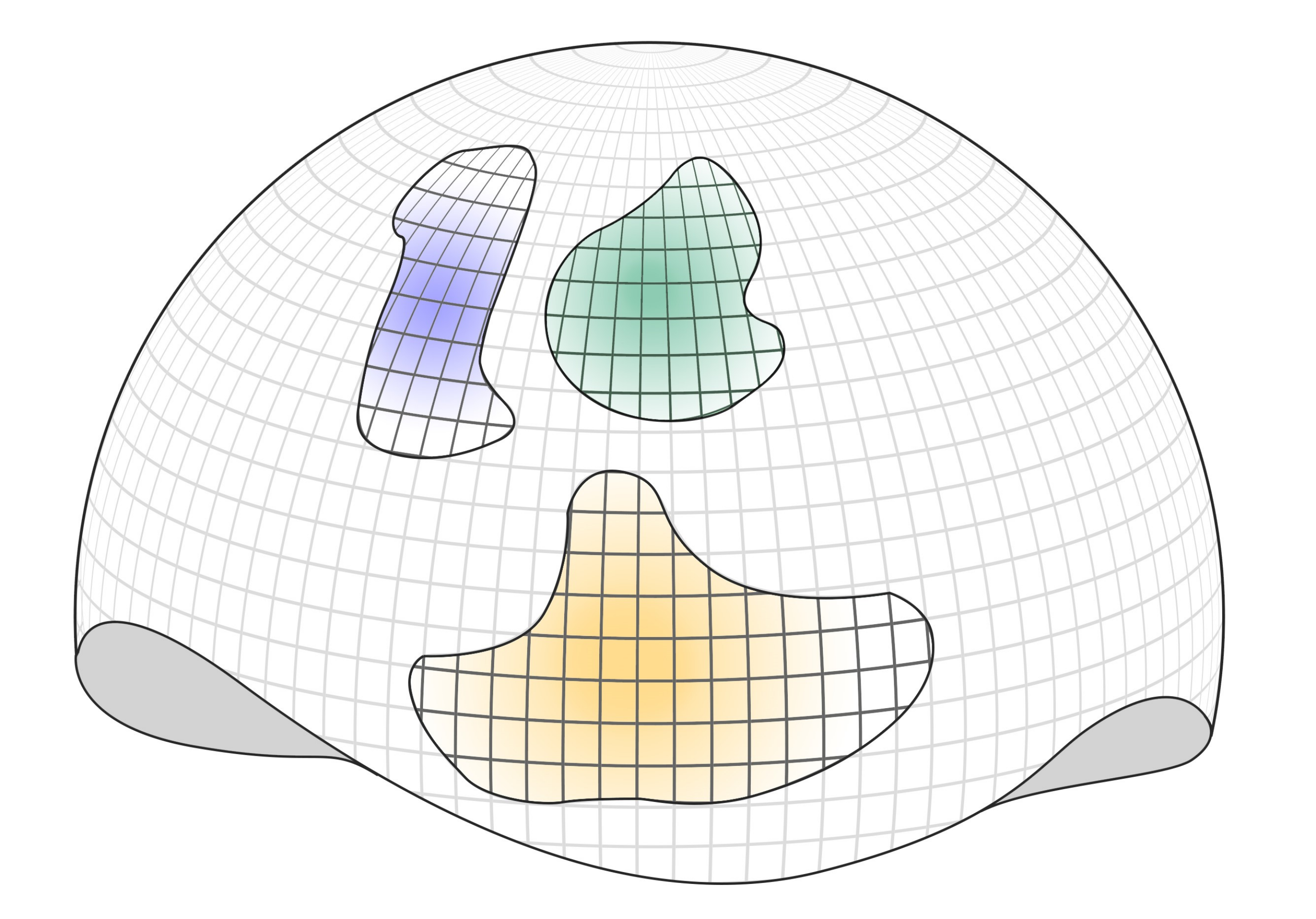}
            \caption{Image classes can be thought of as manifolds in n-dimensional input space.}
         \label{fig:manifolds_in_hyperspace}
     \end{subfigure}
     \hfill
     \begin{subfigure}[b]{0.3\textwidth}
         \centering
         \includegraphics[width=\textwidth]{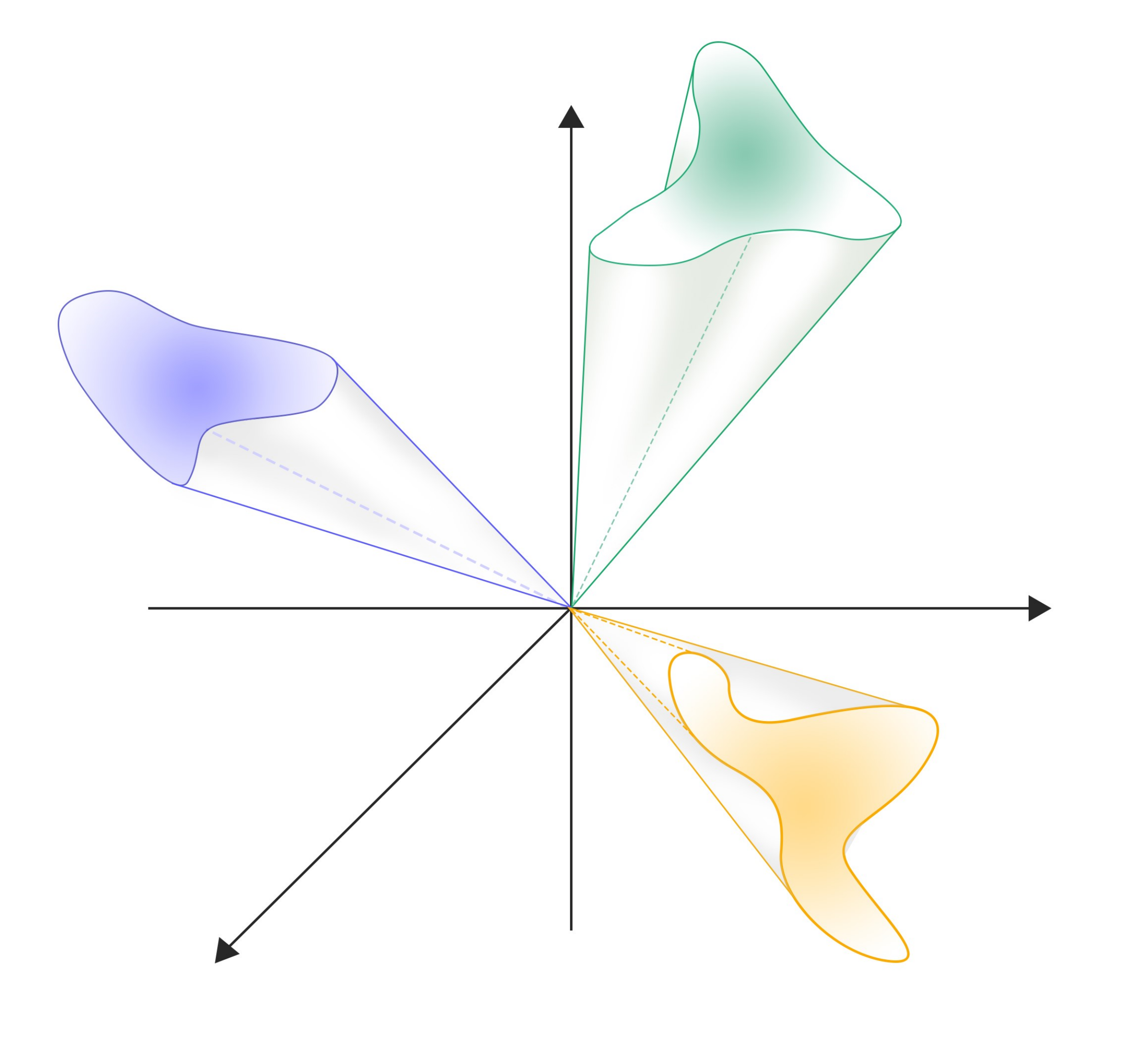}
         \caption{Image classes can be conceptualized as corn-shaped if considering the varying contrast. }
         \label{fig:cone_manifolds_in_n-1_with_scalar_dimension_space}
     \end{subfigure}
     \hfill
     \begin{subfigure}[b]{0.37\textwidth}
         \centering
         \includegraphics[width=\textwidth]{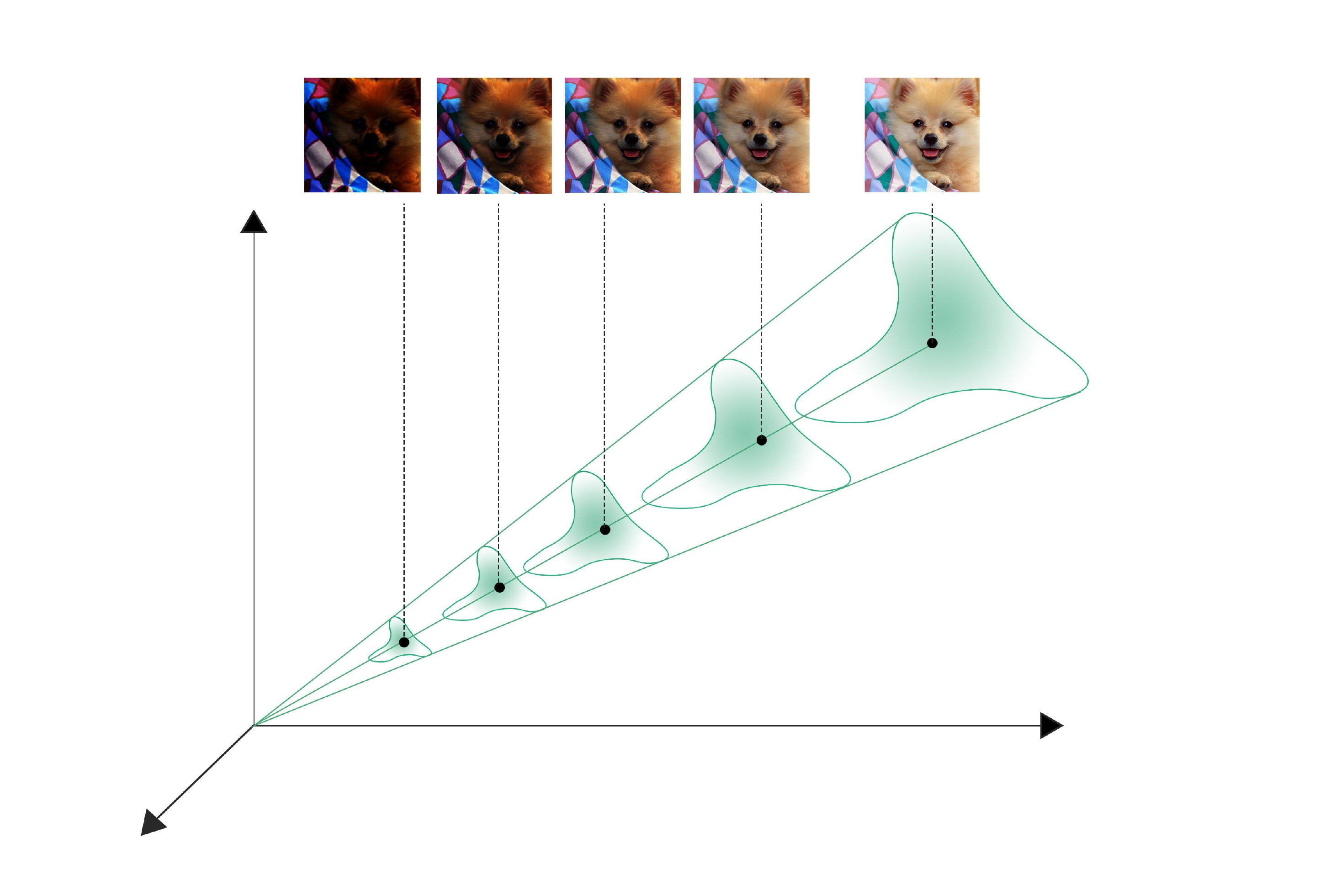}
         \caption{An image specifes a direction in input space, 
          and copies of that image with varying contrast lie along the same direction. }
         \label{fig:cone_disribution_details}
     \end{subfigure}
        \caption{ The directionality (varying contrast) manifests in the intrinsic distribution of images. }
        \label{fig: robustenss_and_input_geometry}
\end{figure}

Using bias terms in neural networks is a common practice. Its theoretical foundation goes back to the invention of artificial neural networks, which are loosely inspired by biological neurons. Biological neurons have some thresholds to determine whether they should "fire” (produce an output that goes to other neurons) \cite{bio_neuro,YANG20201048,HASSABIS2017245}. These thresholds are essentially the same thing as bias terms. From the representation learning perspective, the bias term is widely believed to increase the representational power of neural networks and thus is always needed when designing neural networks to solve a broad array of tasks in computer vision \cite{wang2019bias,MONTAVON20181,Alzubaidi2021ReviewOD}.
In this work, we challenge the commonly-held beliefs of the necessity of including bias terms in neural networks to solve image classification tasks. Our geometric observations suggest the intrinsic distribution of images should incorporate \emph{directionality}, as suggested in Figure \ref{fig: robustenss_and_input_geometry}. With this property holding,  bias terms should not affect models' representational power and performance, even for large modern CNN models such as ResNets \cite{RESNET}. 
Indeed, several recent works like SphereFace~\cite{Liu2017CVPR} and SphereNet~\cite{Liu2017NIPS} achieve strong performance in real-world tasks by  ignoring the bias term and designing angular-inspired losses. Moreover, Hesse et al. \cite{xdnn} report that removing bias terms only has a minor impact on predictive accuracy. Our thorough experimental results also support this argument. 




In addition, we show that neural networks will possess an intriguing property - \emph{scalar (multiplication) invariance} after dropping bias terms. We then extend scalar invariance to CNNs as well as ResNets. This property allows zero-bias networks to perfectly generalize to inputs with different levels of contrast without any data augmentation, which normal neural networks (with biases) usually fail to do so. Based on the scalar invariance property, we further derive more general robustness guarantees that could verify even certain convex regions of input space. In contrast, normal neural networks are highly combinatorial in nature, making such guarantees hardly exist. We also discover that scalar invariant neural networks exhibit complete fairness when predicting zero images. Considering the current issue of bias in AI systems towards certain genders and races \cite{mehrabi2021survey, leavy2018gender}, we believe that scalar invariant neural networks provide a promising solution to eliminate such biases and move towards ethical AI.

We summarize our contributions as follows: (1) We show that the basic building blocks of neural networks are scalar multiplication associative if the bias is ignored. This, in turn, assures the scalar invariant property of convolutional neural networks. By adapting batch normalization-free methods, we can extend scalar invariance to ResNets. We also conduct experiments on a few popular image classification benchmarks to validate the scalar invariant property; (2) Based on the scalar invariant property, we propose two additional robustness properties that verify inputs along certain lines (interpolations) and convex regions of the input space. Empirical validation of the interpolation robustness guarantee is done using image examples from benchmarks such as MNIST and CIFAR-10; (3) We show that scalar invariant neural networks \footnote{We use terms \emph{scalar invariant, zero-bias, without bias} interchangeably to describe the same variant of neural network.} share the same inductive bias as humans, which is a uniform belief in all labels when observing the zero/black images. On the other hand, state-of-the-art models tend to have a strong preference for specific labels; (4) We demonstrate through both empirical results and the theoretical tool known as Neural Tangent Kernel \cite{ntk2018} that zero-bias neural networks and normal neural networks exhibit nearly identical training dynamics; (5) Our geometric observations suggest the intrinsic distribution of images should incorporate \emph{directionality}. Under this property, scalar invariant neural networks should have the same representational power as normal neural networks, thus delivering comparable performances.

\section{Scalar invariant neural networks}
\label{sec:scalar_invariant_neural_networks}

\subsection{Preliminary}
\label{sec:scalar_invariant_neural_networks_preliminary}
A neural network consists of an input layer, hidden layers, and an output layer. For convolutional neural networks, some of the hidden layers are called convolution layers which perform convolution operations on their input tensors with convolution kernels. The outputted tensors are passed to an activation function, commonly \textit{ReLU}, before downsampling through pooling layers. After that, the input tensor is flattened out so that a fully connected network can process it and calculate the final prediction. For classification tasks, the final prediction is represented by a probability distribution over all classes using some activation functions such as \textit{Softmax}. To further investigate the scalar invariant property, we formally denote the input tensor as $X$ and a convolutional neural network as $\mathcal{N}$. Then $\mathcal{N}$ is composed of convolutional layers $\mathcal{F}_i$, pooling layers $\mathcal{P}_i$, and fully connected layers $\mathcal{L}_j$, where $i,j \in \mathbb{N}$. And we denote the final activation function as $\mathcal{A}$ and \textit{ReLU} as $\mathcal{R}$. We think of layers and activation functions as transformations on the input $X$, then the output of the network before the final activation function $\mathcal{A}$ is represented by:
\begin{equation} \label{eqn:computation}
\mathcal{O}(X) = \underbrace{\mathcal{L}_j \circ \mathcal{R} \circ ... \circ \mathcal{R} \circ \mathcal{L}_1}_\text{j Linear layers} \circ \underbrace{ \mathcal{P}_i \circ \mathcal{R} \circ \mathcal{F}_i ... \circ  \mathcal{P}_1 \circ \mathcal{R} \circ \mathcal{F}_1}_\text{i Convolutional layers} \circ X  
\end{equation}


And the final prediction class is determined by the one with the highest probability over all classes $\mathcal{C}$, that is: 
\begin{equation} \label{eqn:argmax}    
\mathcal{N}(X) = \argmax_{ c \in \mathcal{C}}\mathcal{ \{A \circ O}(X)\}
\end{equation}
\subsection{Scalar associative transformations}
We consider the operation inside a convolution layer $\mathcal{F}$ with a kernel $\mathcal{K}$, it is easy to show the associative property with scalar multiplication hold for convolution operations. More formally, let $s$ be a \textbf{positive} scalar s.t. $s \in \mathbb{R}^{+} $, then we have:
\begin{equation}
\label{eqn:associativity1}
\begin{aligned}
\mathcal{ F} \circ (s X) = & \sum_{m}\sum_{n} s X(i+m,j+n)\mathcal{K}(m,n) = s \sum_{m}\sum_{n}  X(i+m,j+n)\mathcal{K}(m,n) = s (\mathcal{F} \circ X)
\end{aligned}
\end{equation}

In addition, the above property also holds for pooling layers $\mathcal{P}$, including max pooling and average pooling. Since both the max and average operation should preserve the scalar multiplication. The same argument also applies to the \textit{ReLU} function. So we have:
\begin{equation}
\label{eqn:associativity2}
\begin{aligned}
\mathcal{ P} \circ (s X) = s (\mathcal{P} \circ X) \textit{ and }  \mathcal{ R} \circ (s X) = s (\mathcal{R} \circ X)
\end{aligned}
\end{equation}
Finally, passing the input $X$ to a fully connected layer $\mathcal{L}$ can be thought of as applying a linear transformation ($\mathcal{W,B}$) on $X$. If we set the bias term $\mathcal{B}$ to $\mathbf{0}$. We will have the scalar associative property. That is:
\begin{equation}\label{eqn:associativity3}
 \mathcal{L} \circ (s X) =  (sX)\mathcal{W}^{T}  = s X\mathcal{W}^{T}   = s (\mathcal{L} \circ X)  
\end{equation}
Note our proofs also use the commutative property which generally holds for matrix and vector multiplications with a scalar. Put together, by setting biases to zeros, we have the scalar (multiplication) associative property holds for the output function, i.e., ($\mathcal{O}(sX) = s\mathcal{O}(X)$).

\subsection{Scalar invariant convolutional neural networks}
Now we consider how to calculate the final prediction of the network $\mathcal{N}$. For classification tasks, the last activation function $\mathcal{A}$ is usually \textit{Softmax}. If we multiply the input $X$ with a scalar $s$ ( $s \in \mathbb{R}^{+} $ ) and pass the product to \textit{Softmax}, it is equivalent to changing the temperature of the distribution. Note that the rank of candidate classes remains the same despite the change in the shape of the distribution. Or in other words, the predicted class by the network $\mathcal{N}$ is scalar (multiplication) invariant:
\begin{equation}
\label{eqn:argmax_consistency1}
\argmax_{c} \frac{e^{s\mathcal{O}(X)_{c}}}{\displaystyle\sum_{c\in\mathcal{C}} e^{s\mathcal{O}(X)_{c}}} = \argmax_{c} \frac{e^{\mathcal{O}(X)_{c}}}{\displaystyle\sum_{c \in \mathcal{C}} e^{\mathcal{O}(X)_{c}}}
\end{equation}
Put together with the scalar associative property of the output function $\mathcal{O}(\cdot)$, we have a scalar invariant neural network:
\begin{equation}
\label{eqn:argmax_consistency2}   
\begin{aligned}
\mathcal{N} (s X) = & \argmax_{c}\mathcal{ \{A} \circ \mathcal{O}(s X)\} = \argmax_{c}\mathcal{ \{A} \circ \mathcal{O}( X)\} =\mathcal{N} ( X)
\end{aligned}
\end{equation}
The concept of scalar invariant neural networks generalizes beyond just convolutional neural networks. In fact, as long as hidden layers perform scalar associative (and commutative) transformations and the last activation function preserves the highest probable candidate under scalar multiplication, the neural network will be scalar invariant. Since an image input $X$ represents a direction in the input space and we have proved that zero-bias neural networks could yield the same prediction along that direction, we could restate this property as directional robustness property.

\begin{lemma} [Directional robustness property] 
\label{lemma: directional_robustness_property}
For any input $X$ to a zero-bias neural network $\mathcal{N}$, the prediction remains the same when $X$ is multiplied by any positive scalar $s$. Formally, we have $\mathcal{N}(sX) = \mathcal{N}(X) \text{ } \text{ } \forall s \in \mathbb{R}^{+}$.
\end{lemma}

\subsection{Scalar invariant ResNet}
We briefly discussed the most simple architecture of convolutional neural networks in the previous section. However, in addition to those basic layers we mention before, modern powerful CNNs also employ extra layers and techniques to address over-fitting and gradient exploding/vanishing issues. For example, ResNet \cite{RESNET} adopts \emph{Dropout} \cite{Dropout}, \emph{Additive Skip Connection} \cite{RESNET} and \emph{Batch Normalization} \cite{Batchnorm} which contributes enormously to its success. First, as dropout layers are disabled during the inference phase, it has no impact on the scalar invariant property. Second, it is trivial to show skip connection is also scalar multiplication associative if the corresponding residual branch $\mathcal{G}$ is also scalar multiplication associative.
\begin{equation}
\label{eqn:factorization}
sX + \mathcal{G}(sX) = s(X + \mathcal{G}(X)) \text{ } \text{ } \forall s \in  \mathbb{R}^{+}
\end{equation}
Lastly, we consider Batch Normalization, which is performed through a normalization transformation that fixes the means and variances of inputs to each layer. Let us use $X_\mathcal{B}$ to denote a mini-batch of the entire training set. Then we have the  batch normalization transformation as follows:
\begin{equation}
\label{eqn:prediction}
\mathcal{BN}(X_{\mathcal{B}}) = \gamma ^{}{ {\hat{X_{\mathcal{B}}}}^{}}+\beta 
\end{equation}
where $\gamma$ and $\beta$ are learnable parameters, and $\hat {X}_{\mathcal{B}}$ is the normalized input, represented by $ {\hat {X}_{\mathcal{B}}}={\frac {X_{\mathcal{B}}^{}-\mu _{\mathcal{B}}^{}}{\sqrt {\left(\sigma _{\mathcal{B}}^{}\right)^{2}+\epsilon }}}$, $\epsilon$ is an arbitrarily small constant. Clearly, we observe that the scalar associative/invariant property doesn't hold for the normalization step, because:




\begin{equation}
\label{eqn:non-bias_matters}    
\gamma {( {sX})}+\beta  =  \gamma {\frac {(sX)^{}-\mu_{\mathcal{B}} ^{}}{\sqrt {\left(\sigma_{\mathcal{B}} ^{}\right)^{2}+\epsilon }}} + \beta \neq s(\gamma {{X}}+\beta)
\end{equation}

Thus, in order to achieve scalar invariance, we can adopt two approaches. Firstly, for small neural networks that do not have severe gradient explosion/vanishing issues, we can drop $\mathcal{BN}$ layers. Secondly, for larger neural networks, we can consider some alternatives to batch normalization. There exists a line of work on exploring efficient residual learning without normalization such as Instance Normalization \cite{instancenormalization}, Fixup \cite{ZhangDM19}, $\mathcal{X}$-DNNs \cite{xdnn}, and NFNets \cite{BrockDS21, BrockDSS21}. The majority of these approaches can be easily adapted to achieve scalar invariance, further information can be found in Appendix \ref{sec: batch_normal}.

\begin{table}[!ht]
\caption{As expected, zero-bias neural networks achieve perfect scalar invariance, while normal neural networks are generally not robust against decreasing the contrast of the input image. Results are replicated thrice and averaged to reduce stochasticity effects, with all variances being below 0.5.}
\centering
\resizebox{\columnwidth}{!}{%
\begin{tabular}{ c|c| c |c c c c c c c c c c c }
\hline
\multicolumn{3}{c|}{}&\multicolumn{11}{c}{Scalar multiplier}\\
\cline{4-14}
\multicolumn{3}{c|}{}&1&0.25&0.15&0.125&0.1&0.075&0.05&0.025&0.01&0.001&0.0001\\ 

\cline{4-14}
\hline
\hline
\multirow{2}{*}{MNIST}&\multirow{2}{*}{FCN}&w/ bias&88.12&87.07&84.46&82.57&79.52&74.76&65.82&42.84&16.34&10.28&10.28\\
&&w/o bias&\textbf{88.27}&\textbf{88.27}&\textbf{88.27}&\textbf{88.27}&\textbf{88.27}&\textbf{88.27}&\textbf{88.27}&\textbf{88.27}&\textbf{88.27}&\textbf{88.27}&\textbf{88.27}\\

\hline
\multirow{2}{*}{Fashion-MNIST}&\multirow{2}{*}{CNN}&w/ bias&\textbf{89.10}&67.10&40.12&32.52&24.16&17.91&12.46&10.12&10.00&10.00&10.00\\
&&w/o bias&89.02&\textbf{89.02}&\textbf{89.02}&\textbf{89.02}&\textbf{89.02}&\textbf{89.02}&\textbf{89.02}&\textbf{89.02}&\textbf{89.02}&\textbf{89.02}&\textbf{89.02}\\


\hline
\multirow{2}{*}{CIFAR-100}&\multirow{2}{*}{ResNet18}&w/ bias&\textbf{67.62}&19.86&8.20&6.11&4.16&2.58&1.69&1.06&1.01&1.01&1.01\\
&&w/o bias&67.33&\textbf{67.33}&\textbf{67.33}&\textbf{67.33}&\textbf{67.33}&\textbf{67.33}&\textbf{67.33}&\textbf{67.33}&\textbf{67.33}&\textbf{67.33}&\textbf{67.33}\\

\hline
\multirow{2}{*}{ImageNet \cite{imagenet}}&\multirow{2}{*}{ResNet50}&w/ bias&\textbf{75.37}&66.72&57.84&53.62&47.27&37.61&21.81&3.39&0.21&0.10&0.10\\
&&w/o bias&73.82&\textbf{73.82}&\textbf{73.82}&\textbf{73.82}&\textbf{73.82}&\textbf{73.82}&\textbf{73.82}&\textbf{73.82}&\textbf{73.82}&\textbf{73.82}&\textbf{73.82}\\


\hline
\end{tabular}
}


\label{tab:scalar_inv_tab}
\end{table}

\subsection{Scalar invariance evaluation}
\label{sec:scalar_inv_eval}

In this section, we conduct a series of experiments to verify the scalar invariance property of zero-bias neural networks and their normally trained counterparts. We train both types of neural networks using the same configuration, except for the option of using bias, on several popular image classification benchmarks. More training details can be founded in Appendix \ref{sec: training_details}. We further demonstrate the effect of scalar invariance by evaluating their accuracy on test sets multiplied by different scalars, ranging from $1$ to $0.0001$. The results, which are presented in Table \ref{tab:scalar_inv_tab}, suggest that zero-bias networks and normal networks achieve similar accuracies when the scalar is set to $1$. However, when the contrast/scalar multiplier of the input image decreases, normal networks show a lack of robustness as their accuracy declines at varying rates. In contrast, zero-bias networks achieve scalar invariance as expected, and their performance remains unchanged regardless of the varying contrast of input images. We also train both types of models using augmented training sets that involve multiplication of the scalars used in test evaluation. We find that with-bias models trained on augmented data still perform poorly when the scalar multiplier is extremely small, such as $0.001$ and $0.0001$. For larger scalar ranges from $0.25$ to $0.01$, with-bias models are merely comparable to zero-bias models. These results demonstrate a significant advantage of zero-bias networks in terms of data efficiency.

\section{Interesting robustness properties}
\label{sec:robfair}

\begin{figure}
     \centering
     \begin{subfigure}[b]{0.44\textwidth}
         \centering
            \includegraphics[width=\textwidth]{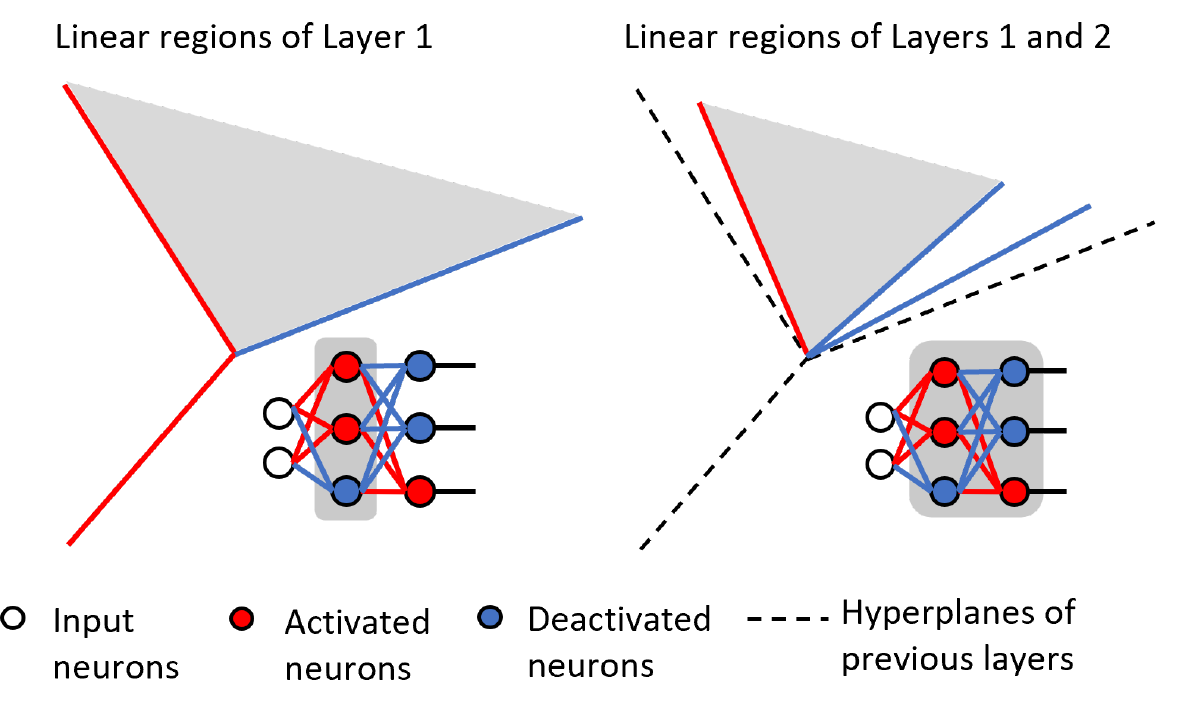}
            \caption{Core-shape unbounded regions of zero-bias neural networks are formed by hyper-planes that pass through the origin. }
         \label{fig:zerobias_linear_regions}
     \end{subfigure}
     \hfill
     \begin{subfigure}[b]{0.25\textwidth}
         \centering
         \includegraphics[width=\textwidth]{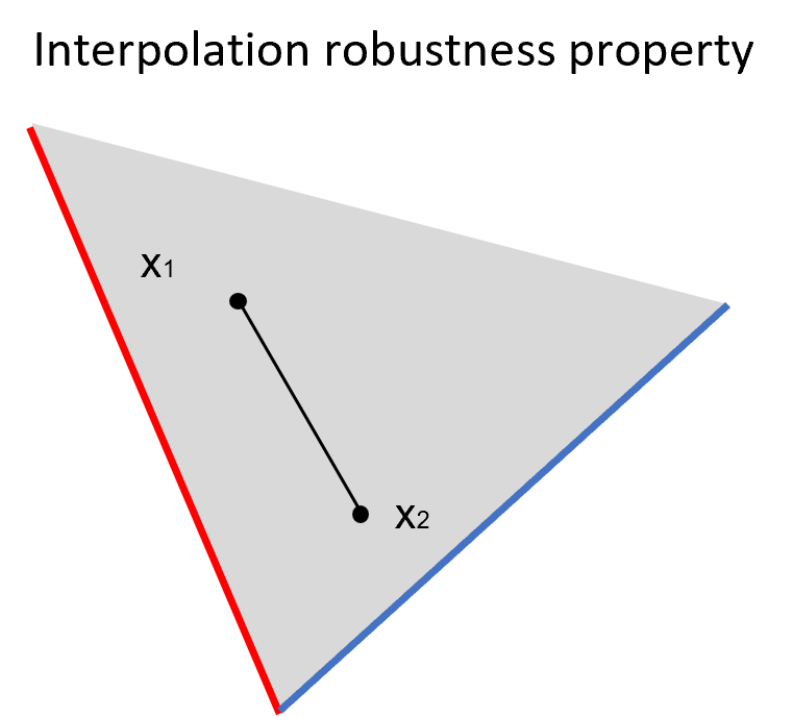}
         \caption{Any point lying on the line between $x_1$ and $x_2$ should yield the same prediction as $x_1$ and $x_2$.}
         \label{fig:Interpolation_robustness_property}
     \end{subfigure}
     \hfill
     \begin{subfigure}[b]{0.27\textwidth}
         \centering
         \includegraphics[width=\textwidth]{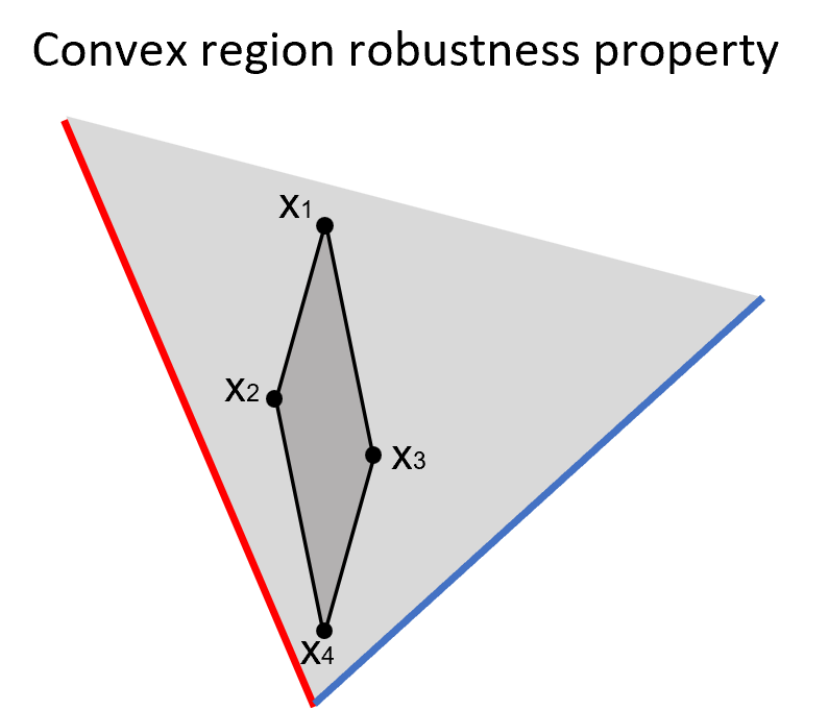}
         \caption{Any point lying inside the convex region formed by inputs ${x_i}$ should yield the same prediction as $x_i$.}
         \label{fig:Convex_region_robustness_property}
     \end{subfigure}
        \caption{Zero-bias neural networks exhibit unique robustness properties when inputs share the same neural activation pattern and are predicted identically by the neural network. }
        \label{fig: Robustenss_and_input_geometry.}
\end{figure}

Despite achieving remarkable success in a wide range of tasks, neural networks have been proven not robust under even small perturbations to the input \cite{nonrobust,ThreatAd}, which accelerates the study of neural network verification and attacks. We find that zero-bias networks exhibit some interesting robustness guarantees that are rarely identified in common neural networks. 

Given that these guarantees of robustness are closely tied to specific regions within the input space, it is pertinent to explore how zero-bias neural networks divide the geometry of the input space. When the bias terms are eliminated, the hyperplanes defined by each neuron will originate from the origin. When these hyperplane arrange together, they create multiple core-shaped unbounded regions that differ from the typical convex regions formed by normal neural networks, as illustrated in Figure \ref{fig:zerobias_linear_regions}. To better illustrate the interesting robustness properties of zero-bias networks, we first introduce the notion of neural activation patterns \cite{geng2022}.




\begin{definition}[Neural Activation Pattern] A \emph{Neural Activation Pattern (NAP)} of a neural network $\cN$ is a tuple $\cP_x \mathbin{:=} (A, D)$, where $A$ and $D$ are collections of all activated and deactivated neurons respectively when passing $x$ through $\cN$.\end{definition}

\begin{theorem}[Interpolation robustness property]
\label{lemma:interpolation_robustness_property}
For any two inputs $X_1$ and $X_2$ that have the same prediction and neural activation pattern by network $\mathcal{N}$, i.e., $\mathcal{N}(X_1) = \mathcal{N}(X_2)$ and $\cP_{X_1} = \cP_{X_2}$, their linear interpolation also yield the same prediction, that is, $\mathcal{N}(\lambda X_1 + (1-\lambda)X_2) = \mathcal{N}(X_1) = \mathcal{N}(X_2)$, where $\lambda \in [0,1]$.
\end{theorem}

Assuming that two points share the same prediction and neural activation pattern, it can be proven that their interpolation will also share the same prediction and neural activation pattern. Please refer to Appendix \ref{sec: proof_interpolation_robustness_property} for detailed proof. What's even more interesting is that this property can be extended to the multiple inputs setting, where a convex region can provide robustness assurance.

\begin{theorem}
[Convex region robustness property]
Let $\{X_i \mid i \in \{1,2,\dots,n\}\}$ be a collection of inputs that have the same prediction and neural activation pattern by network $\mathcal{N}$, we denote the convex polygon formed by vertices ${X_i}$ as $\mathcal{M}$. Then, for any point $m$ that lies inside the polygon $\mathcal{M}$, $m$ also yield the same prediction as $X_i$, that is, $ \mathcal{N}(m) = \mathcal{N}(X_i) \text{  
  } \text{ } \forall m \in \mathcal{M}  \text{ }  \forall i \in \{1,2,...,n\} $. 
\end{theorem}
As $m$ can always be represented by some linear combination of vertices $\{X_i\}$, the convex region robustness property holds as the direct result of \textbf{Theorem} \ref{lemma:interpolation_robustness_property}. In contrast, such guarantees hardly exist on normal neural networks due to their highly combinatorial nature. Furthermore, recent research has shown that ignoring bias can enhance the robustness of models, as demonstrated in \cite{MPTs,MPTs_wh}.

\begin{figure}[ht]
     \centering
         \includegraphics[width=\textwidth]{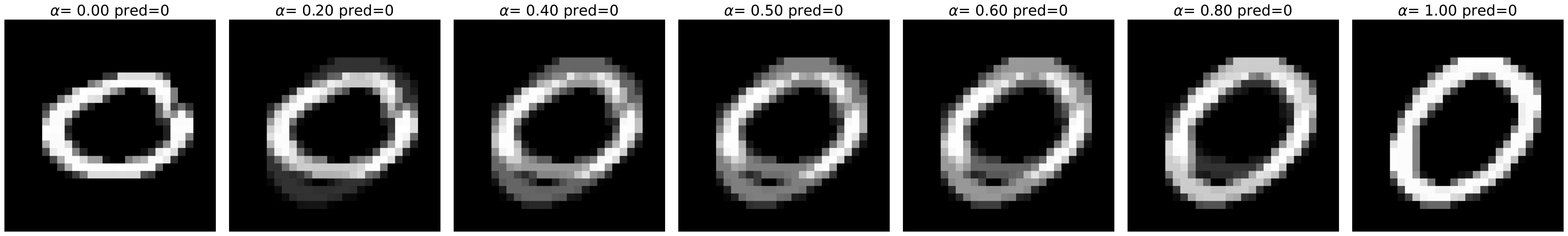}
         \label{fig:Interp_mnist_ex1}
         \centering
         \includegraphics[width=\textwidth]{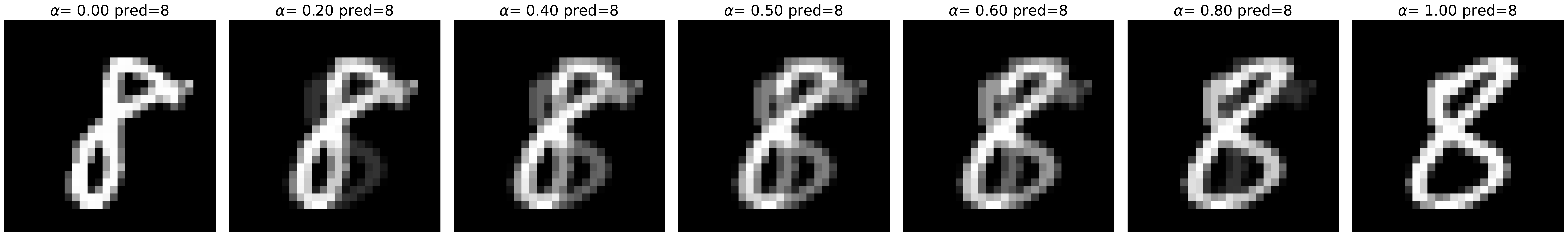}
         \label{fig:Interp_mnist_ex2}
         \centering
         \includegraphics[width=\textwidth]{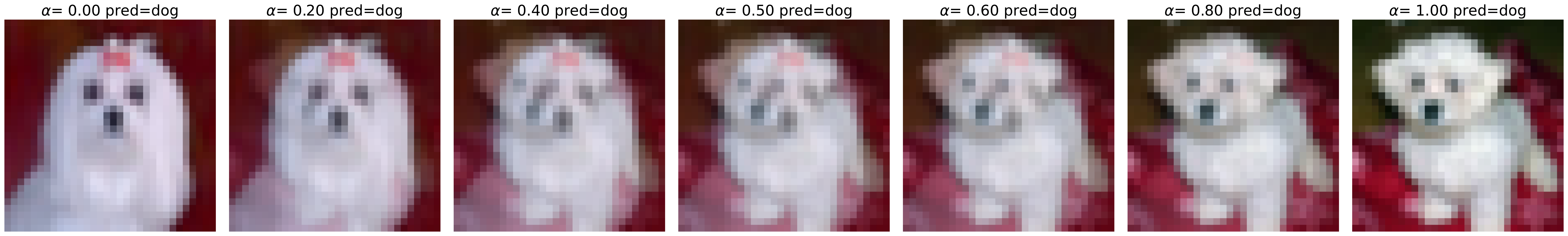}
         \label{fig:Interp_cifar10_ex1}
         \centering
         \includegraphics[width=\textwidth]{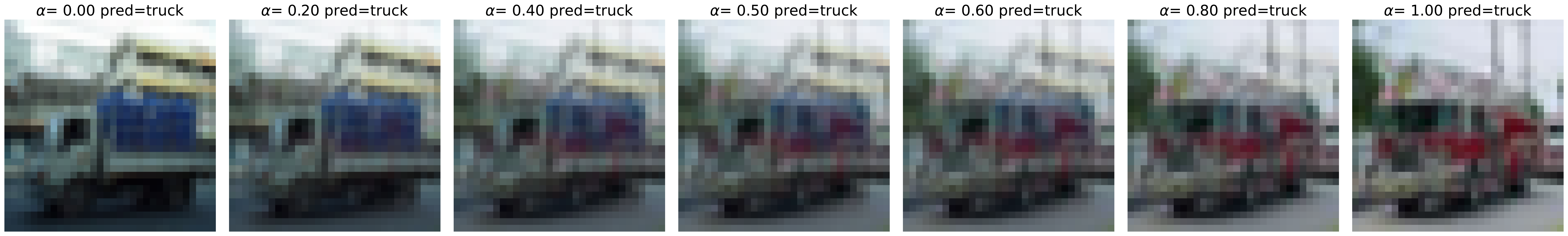}
         \label{fig:Interp_cifar10_ex2} 
        \caption{ The left-most and right-most images are from the original MNIST (the first two rows) and CIFAR10 (the last two rows), whereas synthesized/interpolated images are in the middle. We find that any interpolated images will yield the same prediction as a result of the interpolation robustness property (\textbf{Theorem} \ref{lemma:interpolation_robustness_property}). Please refer to Appendix \ref{sec: interpolation_robustness_examples} for additional examples.}
        \label{fig:Interpolation}
\end{figure}

To test the interpolation robustness property, we conducted experiments using visual examples sourced from MNIST and CIFAR-10. Following neural network training, we search for image pairs that shared the same prediction and neural activation pattern. We then interpolate 1000 images between each pair and confirm that each interpolation yields the same prediction, as expected. Figure \ref{fig:Interpolation} presents some examples of our findings. 

However, it is important to note that such robustness guarantees are rarely identified in larger and more accurate neural networks. For example, with the small neural networks used in our experiments comprising only 30 neurons and achieving accuracies of 32.27\% and 29.6\% on MNIST and CIFAR-10, respectively, we can easily identify many qualified pairs. In larger networks with accuracies of around 80\%, we still find a few qualified pairs. However, in even larger networks with accuracies of over 90\%, we are unable to find any examples of the interpolation robustness property. This is due to the fact that as the number of neurons grows, the input space becomes more scattered, reducing the likelihood of two or more inputs sharing the same neural activation pattern. While this seems like a new \emph{No Free Lunch Theorem} in terms of the trade-off between interpolation robustness and accuracy, we believe there are methods to improve model accuracy while still maintaining these robustness guarantees. For example, we could design new training objectives to control the diminishing margin between hyperplanes of networks. We leave this as a direction for future work.

\section{Fairness on the zero image}
\label{sec:fair}

\begin{figure*}[t]
     \centering
     \begin{subfigure}[t]{0.32\textwidth}
         \centering
         \includegraphics[width=\textwidth]{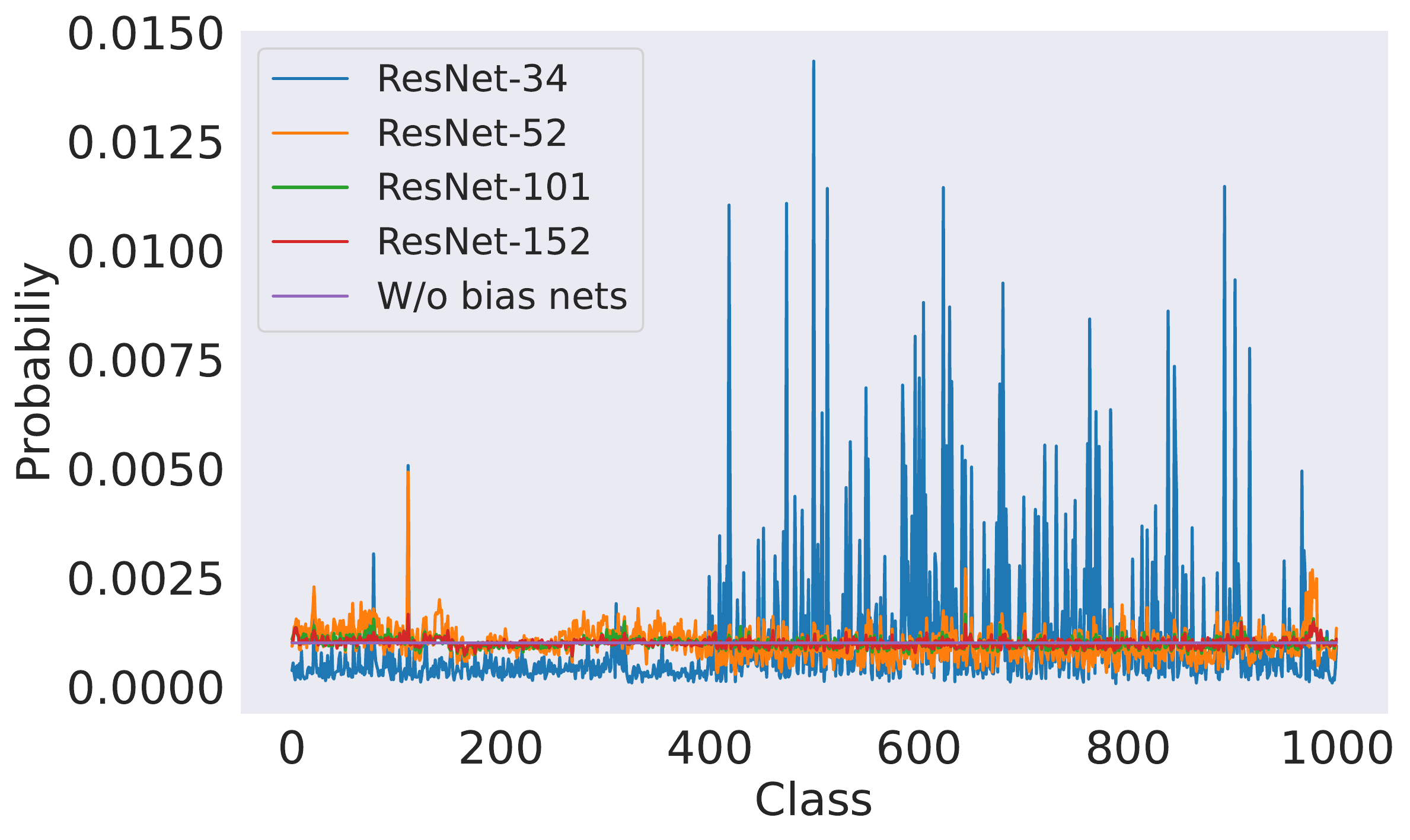}
         \caption{The predicted probability of ResNets on the zero image.}
         \label{fig:ResNet_bias}
     \end{subfigure}
     \hfill    
      \begin{subfigure}[t]{0.32\textwidth}
         \centering
         \includegraphics[width=\textwidth]{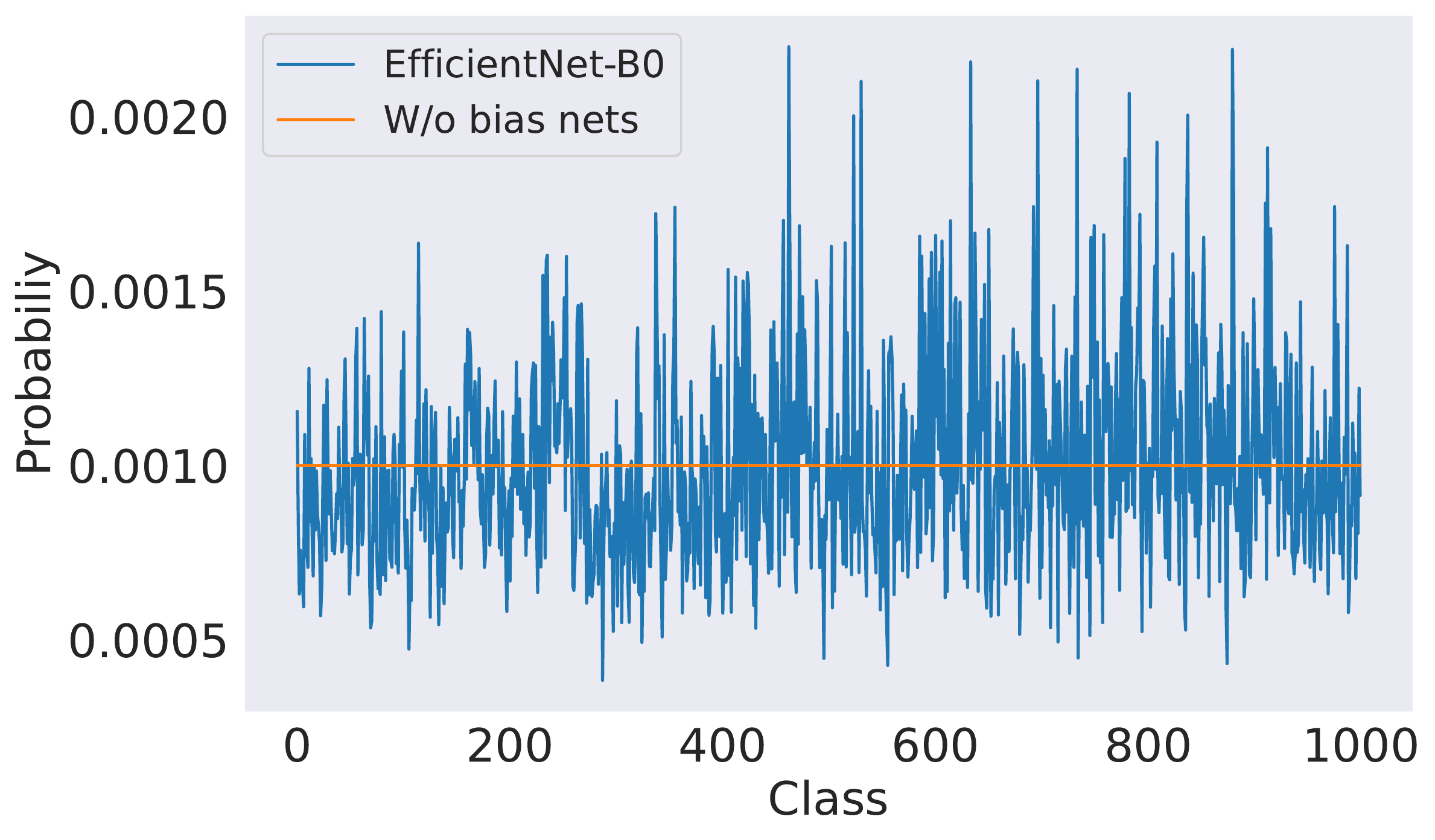}
         \caption{The predicted probability of EfficientNet on the zero image.}
         \label{fig:EfficientNet_bias}
      \end{subfigure}
     \hfill    
      \begin{subfigure}[t]{0.32\textwidth}
         \centering
       \includegraphics[width=\textwidth]{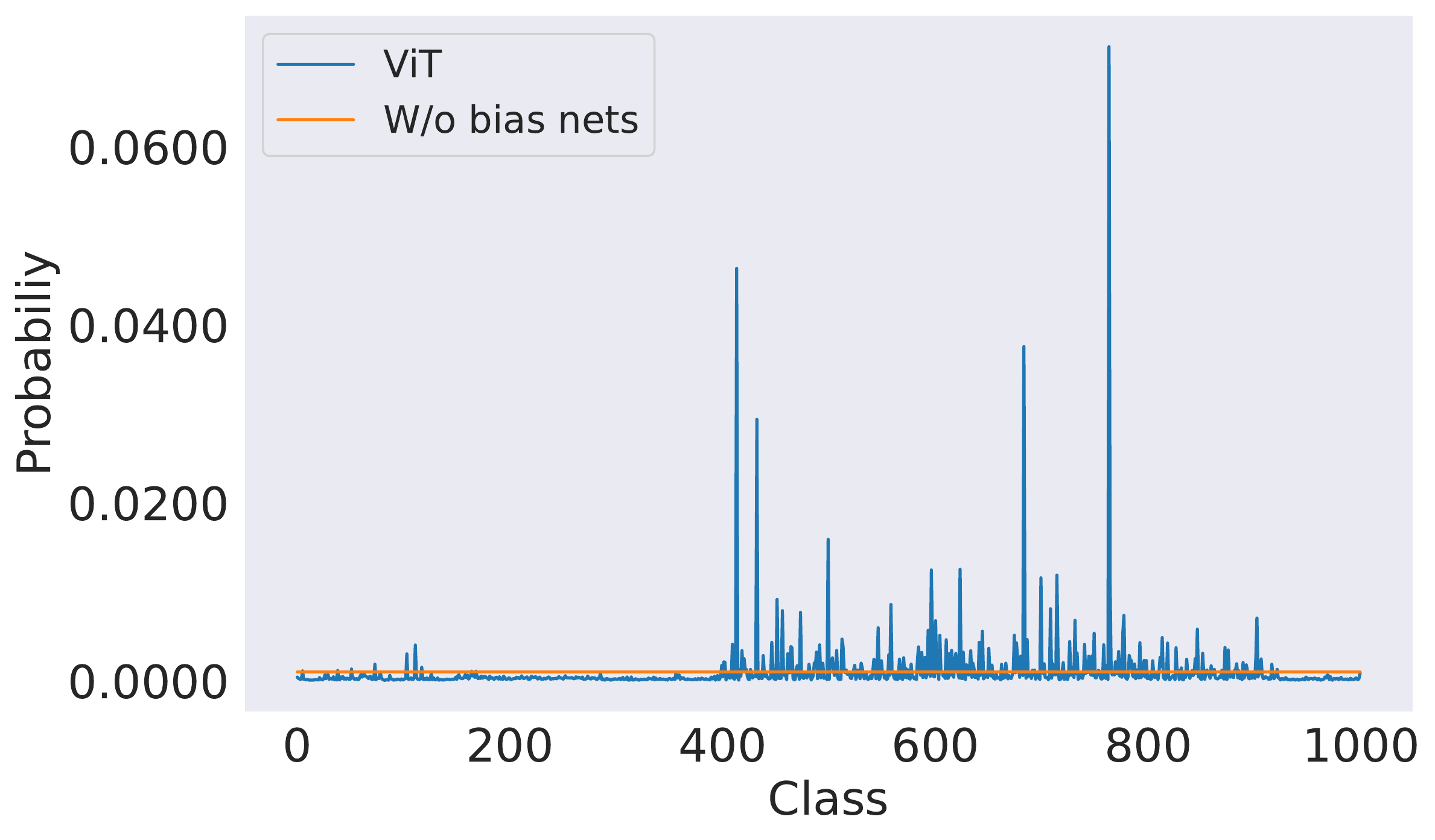}
         \caption{The predicted probability of ViT on the zero image.}
         \label{fig:ViTt_bias}
      \end{subfigure}      
        \caption{SOTA models are biased towards to certain classes when predicting the zero image, whereas scalar invariant neural networks are unbiased, i.e, having a uniform belief in all classes like humans. }
        \label{fig:Bias_at_zero}
\end{figure*}
This section investigates the fairness of models when predicting the zero image, which refers to an image where all pixel values are set to zero. From a human perspective, the zero image contains no discernible information, resulting in maximum information entropy. As such, it is equally likely for the zero image to belong to any class, meaning that it follows a uniform distribution. It is easy to demonstrate that scalar invariant neural networks possess the same inductive bias as humans, since:
\begin{equation}
    \label{eqn:fireness_of_zero_image}
    \begin{aligned}
\mathcal{N}(\mathbf{0}) = & \argmax_{c }\mathcal{ \{A \circ \mathcal{O}(\mathbf{0})\}} = \argmax_{c } \frac{e^{\mathbf{0}}}{\displaystyle\sum_{c\in\mathcal{C}} e^{\mathbf{0}}} = \argmax_{c} \frac{1}{|\mathcal{C}|}
    \end{aligned}
\end{equation}
Nevertheless, this may not hold for normal neural networks with bias terms, even those that are considered state-of-the-art models, as they may exhibit bias towards certain classes. Figure \ref{fig:Bias_at_zero} presents selective results on models' bias/fairness when predicting the zero image. It is noteworthy that all selected models display some level of bias. Specifically, among the three state-of-the-art models, EfficientNet \cite{EfficientNet} shows less bias, while ViT \cite{ViT} is heavily biased towards certain classes.

However, this is a significant concern when applying AI to real-world applications such as gender classification, as current AI systems have shown to have problems with bias, including gender and racial bias \cite{ShamARKAOOA23, buolamwini_2019}. We believe that this issue is deeply rooted in the inductive bias of normal neural networks and may not be easily addressed by using augmented data or changing training objectives. In contrast, zero-bias neural networks exhibit a uniform belief in all potential classes, which cannot be altered even by training with imbalanced data, as it serves as an inductive bias of the model. While our study focuses on the fairness of zero-bias neural networks in the context of the zero image, we believe that this fairness property has the potential to be extended to other scenarios. For instance, it may help address gender or racial bias, providing a promising path towards achieving ethical AI. We intend to explore this direction in our future work.

\section{Training dynamics and expressiveness }
\label{sec: training dynamics}

\begin{figure*}[h!]
     \centering
     \begin{subfigure}[t]{0.2435\textwidth}
         \centering
         \includegraphics[width=\textwidth]{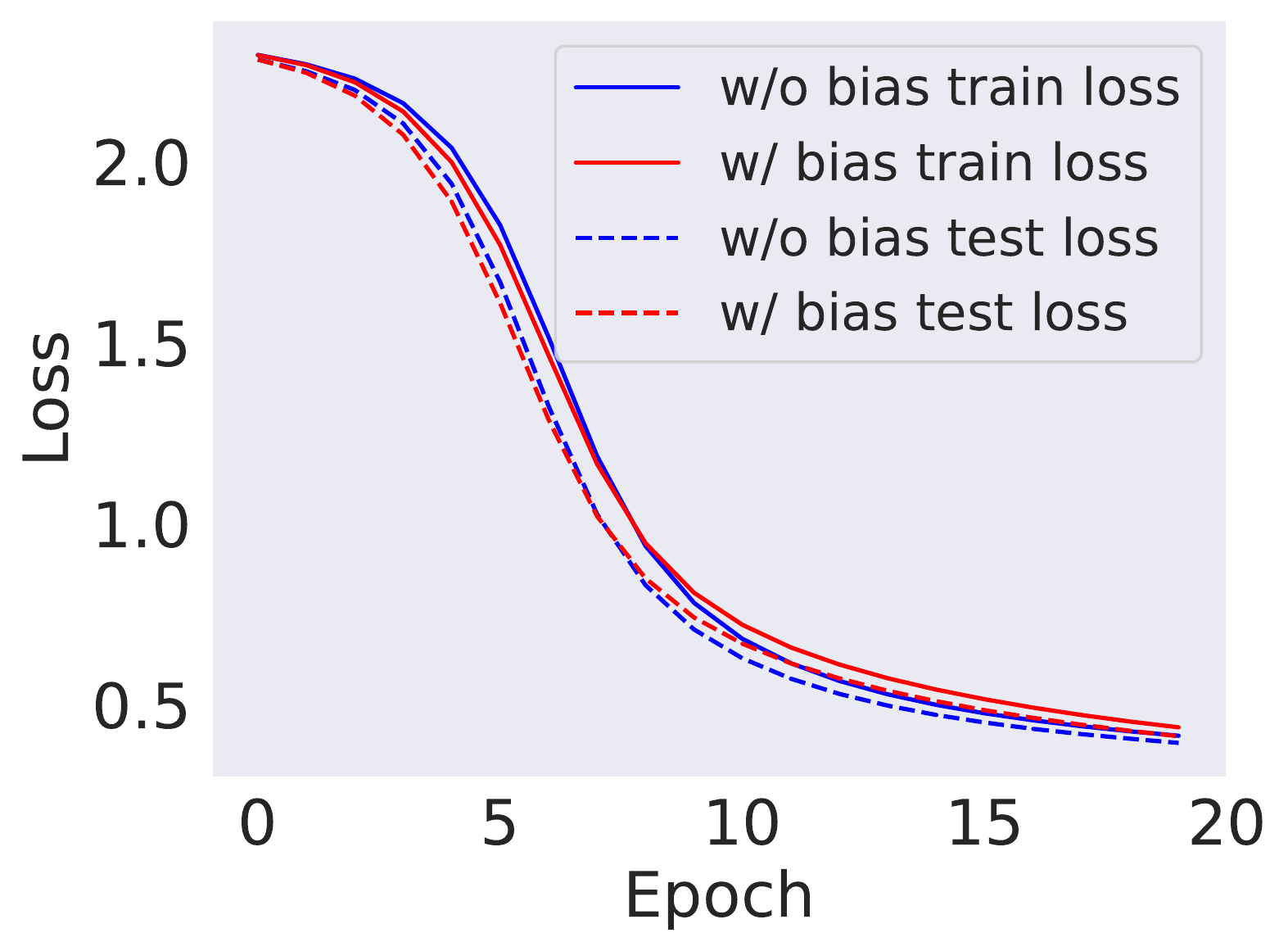}
         \caption{Loss curves of FCNs on MNIST.}
         \label{fig:FCN_loss}
     \end{subfigure}
     \hfill
     \begin{subfigure}[t]{0.2435\textwidth}
         \centering
         \includegraphics[width=\textwidth]{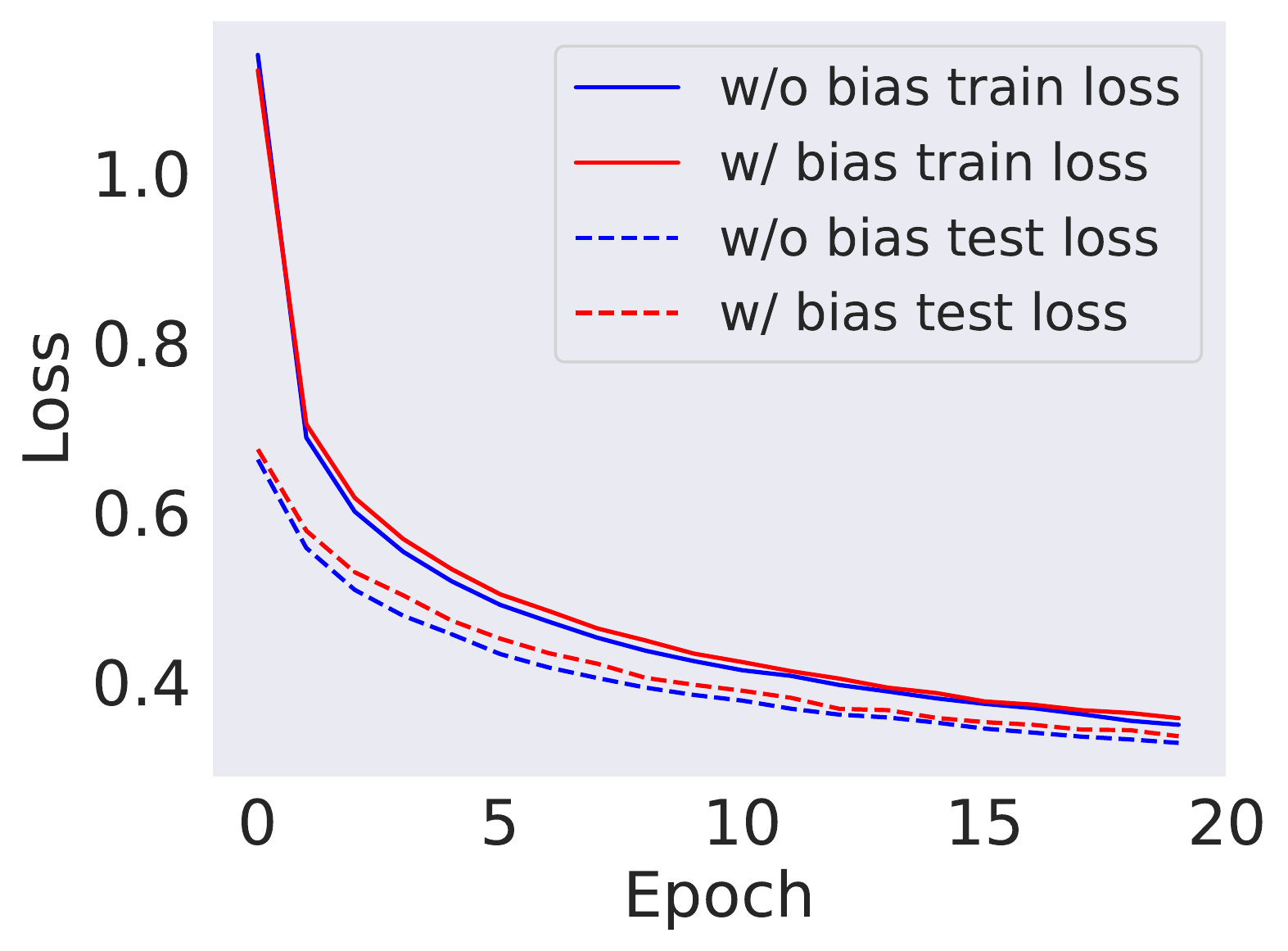}
         \caption{Loss curves of CNNs on Fashion-MNIST.}
         \label{fig:CNN_loss}
     \end{subfigure}
     \hfill
      \begin{subfigure}[t]{0.239\textwidth}
         \centering
         \includegraphics[width=\textwidth]{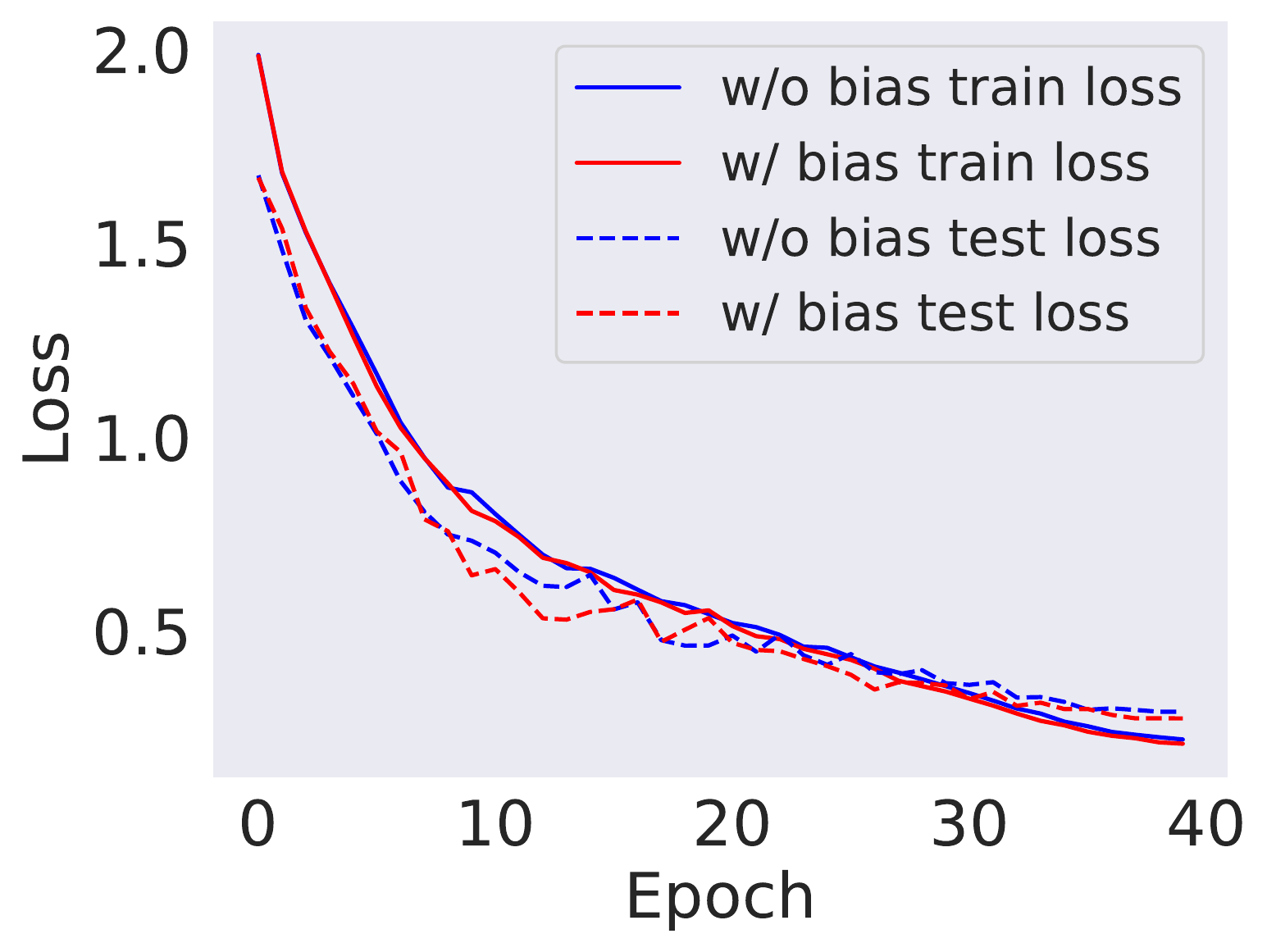}
         \caption{Loss curves of ResNet9 on CIFAR-10.}
         \label{fig:Imagenette_loss}
      \end{subfigure}
     \hfill
    \begin{subfigure}[t]{0.239\textwidth}
         \centering
         \includegraphics[width=\textwidth]{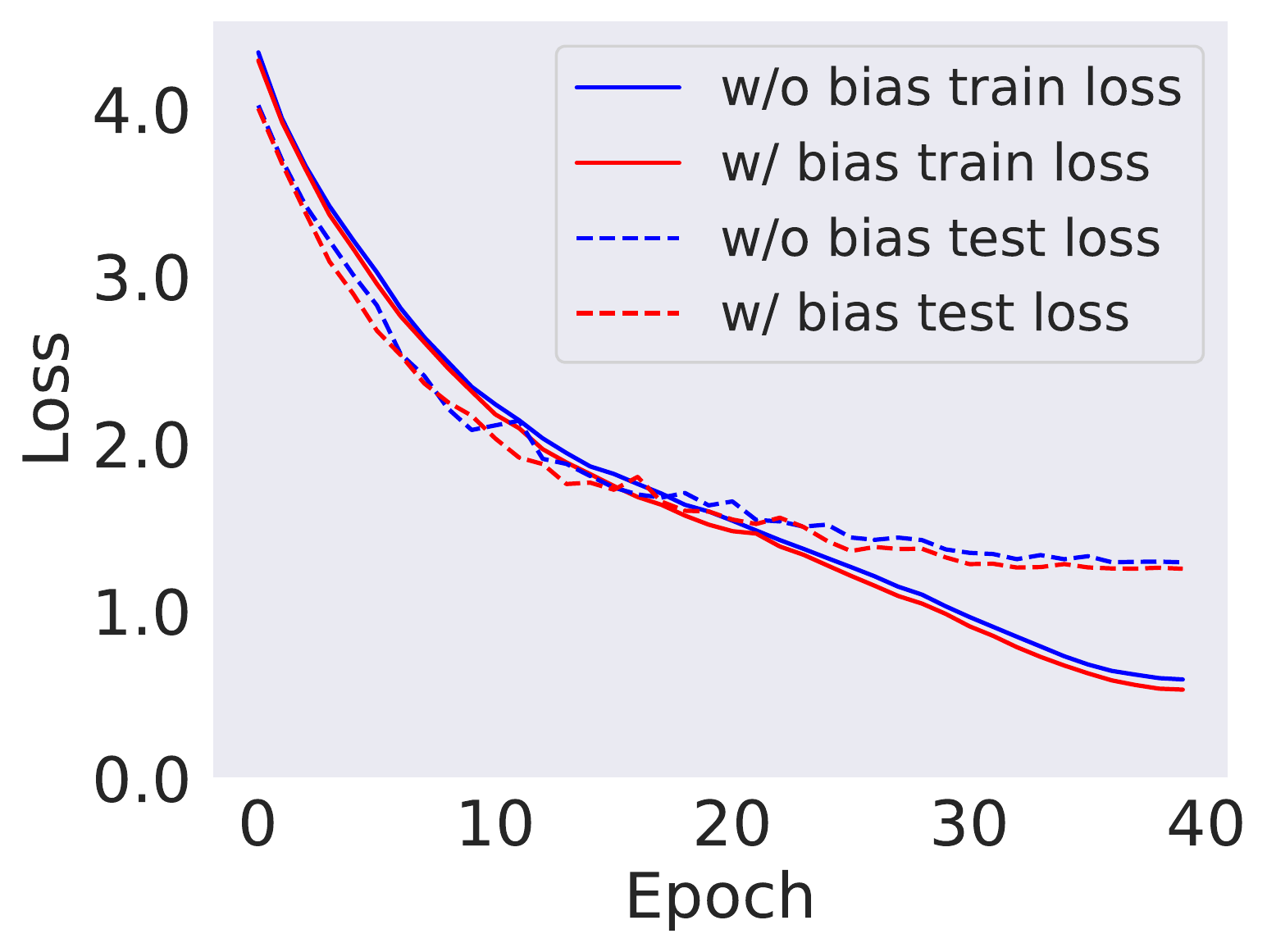}
         \caption{Loss curves of ResNet18 on CIFAR-100.}
         \label{fig:CIFAR100_loss}
     \end{subfigure}
        \caption{Loss curves of normally trained neural networks, and their scalar invariant
counterparts are almost identical, which supports our argument that removing bias doesn't impact the training dynamics and generalization of models on image classification tasks. 
}
        \label{fig:Loss_curve}
\end{figure*}





\subsection{Training dynamics}
\label{sec: remarks}
In this section, we demonstrate that removing the bias terms does not affect the training dynamics of neural networks. We begin by demonstrating our observation through the use of a theoretical tool known as the Neural Tangent Kernel (NTK). The NTK is a  kernel that explains how neural networks evolve during training through gradient descent \cite{ntk2018}. It provides valuable insights into why sufficiently wide neural networks can converge to a global minimum when trained to minimize an empirical loss. We extend the two key results from the original paper to zero-bias cases, as follows:

\begin{remark}
When the width of networks goes to infinite, both NTKs of zero-bias neural networks and normal neural networks converge in probability to the same deterministic limit.

\end{remark}

\begin{remark}
In the infinite-width limit, both NTKs of zero-bias neural networks and normal neural networks stay asymptotically the same constant during training.

\end{remark}

To summarize, from the perspective of NTK formulation, the bias terms do not have an impact on the training dynamics of models. We provide further details for supporting these two remarks in the Appendix. \ref{sec: key_results}. While successful neural networks are not typically operated in the kernel regime, the Neural Tangent Kernel provides valuable insights that bias terms do not play a key role in the training dynamics of neural networks. In addition, we empirically evaluate and compare both zero-bias networks and their normally trained counterparts on some popular image classification benchmarks, as shown in  Figure \ref{fig:Loss_curve}. Highly overlapped training loss curves indicate that both types of models have almost identical training dynamics, which aligns with our analysis results using NTK. More surprisingly, the two different types of models also exhibit very similar generalization capabilities on unseen datasets. We also aim to dive deeper into the generalization behavior of zero-bias networks  in future work.



\subsection{Geometric insights on expressiveness}
\label{sec:geo_insights}

\begin{figure}[ht]
\centering
     \begin{subfigure}[t]{0.24\textwidth}
         \includegraphics[width=\textwidth]{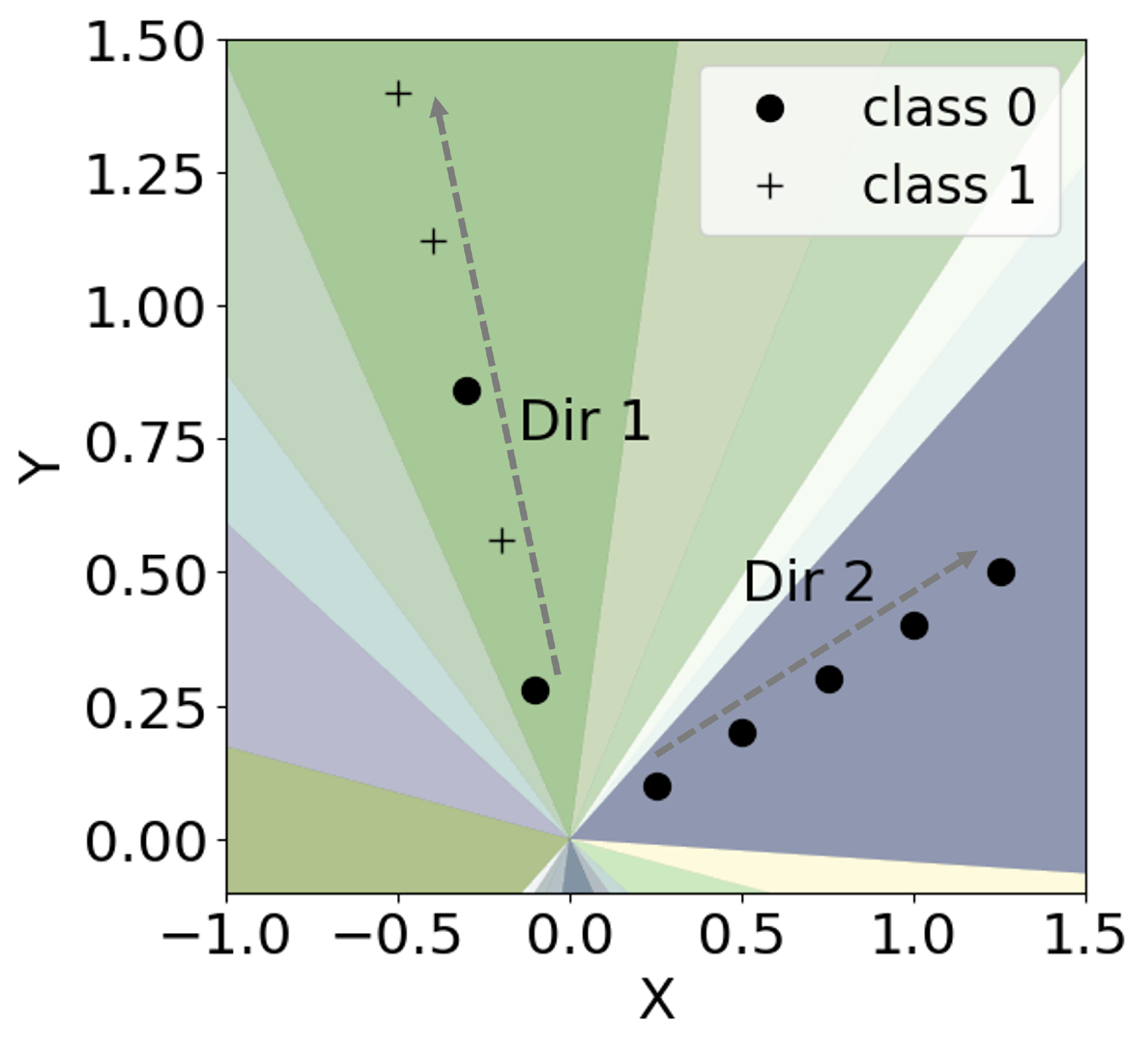}
         \caption{ Dir1 and Dir2 stay within unbounded regions of the zero-bias network.} 
         \label{fig:linear_regions_nb}
     \end{subfigure}
     \begin{subfigure}[t]{0.24\textwidth}
         \includegraphics[width=\textwidth]{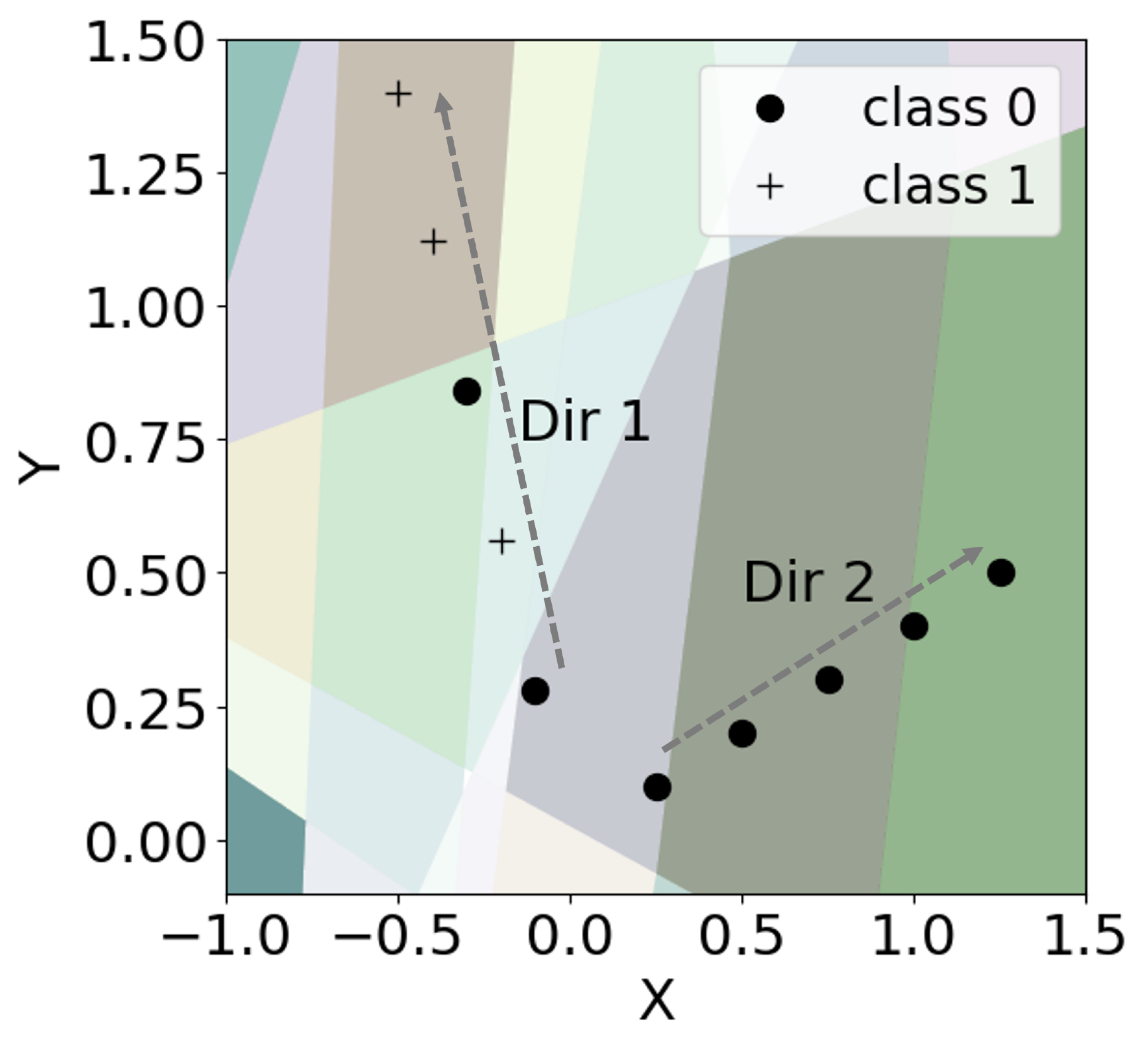}
         \caption{ Dir1 and Dir2 traverse multiple convex regions of the with-bias network.}
         \label{fig:linear_regions_wb}
         \hfill
     \end{subfigure}
     \begin{subfigure}[t]{0.25\textwidth}
         \includegraphics[width=\textwidth]{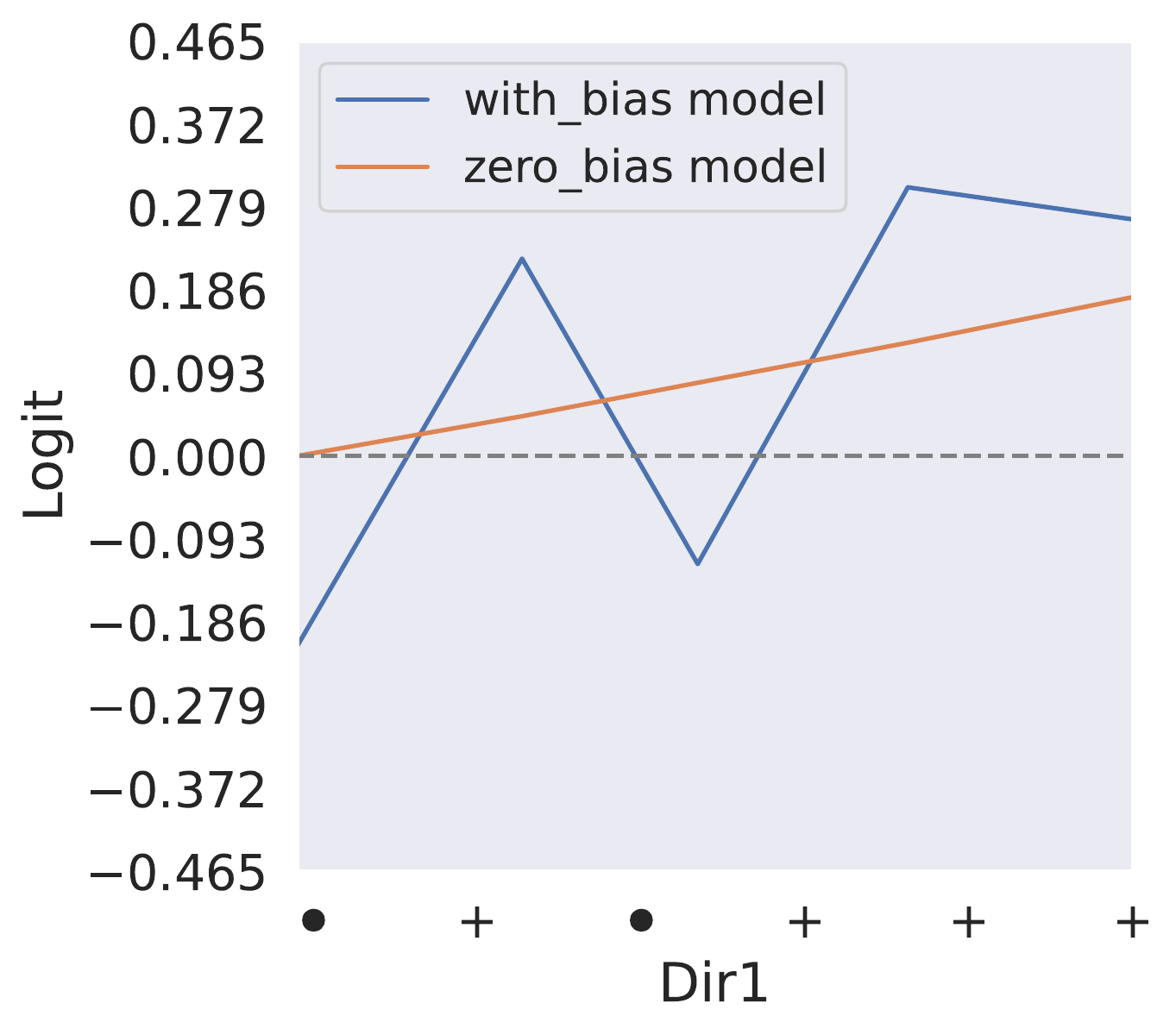}
         \caption{Normal networks can fit data in Dir1 whereas zero-bias networks fail to do so.} 
         \label{fig:logit_dir1}
     \end{subfigure}
     \begin{subfigure}[t]{0.25\textwidth}
         \includegraphics[width=\textwidth]{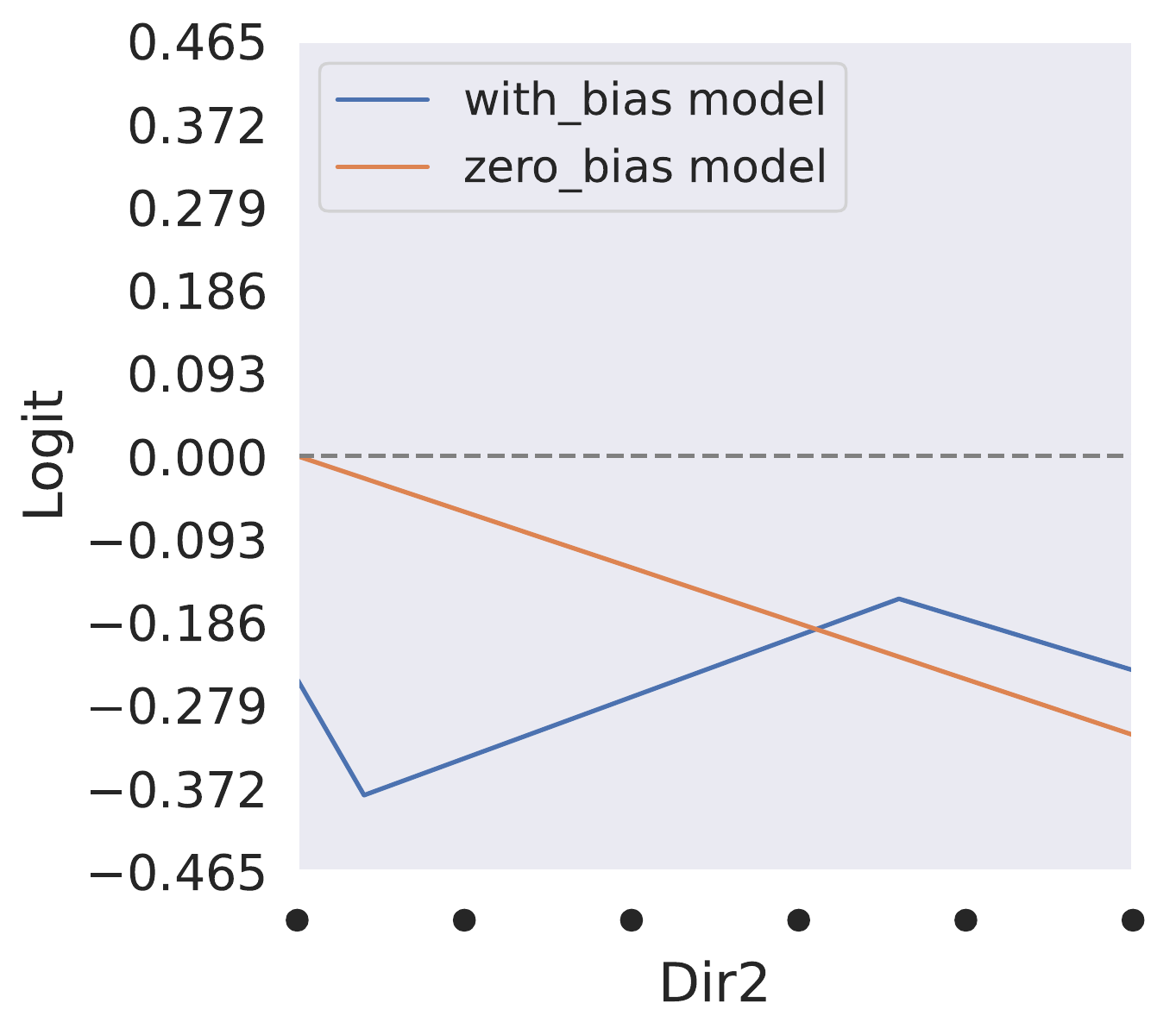}
         \caption{Both normal and zero-bias networks can fit data in Dir2 (directionality holds).}
         \label{fig:logit_dir2}
         \hfill
     \end{subfigure}

        \caption{The first two subfigures show \emph{Dir} 1 and \emph{Dir} 2 on the input space for both zero-bias networks and normal networks, respectively. The latter two subfigures represent the prediction logit function of zero-bias networks and normal networks along \emph{Dir} 1 and \emph{Dir} 2, respectively.}
        
        \label{fig:linear_reg}
          
\end{figure}

It is a widely held belief that eliminating bias from neural networks can diminish their representational power, ultimately affecting the accuracy of models. For instance, Xu et al. \cite{xu2020neural} show that neural networks linearize outside of their training regime once omitting the biases. To this end, we provide geometric insights to show that zero-bias networks are comparable with normal networks in expressive capabilities when solving image classification tasks. 

Since a neural network can be thought of as a piece-wise (linear) function defined over many convex polytopes \cite{LR1,LR2}, we plot linear regions of a simple 3-layer neural network and its zero-bias counterpart on a simple 2D input space to study their representational power in Figure \ref{fig:linear_reg}. Our aim is to illustrate how these networks perform on two simple binary classification tasks characterized by \emph{Dir (Direction)} 1 and \emph{Dir} 2. In \emph{Dir} 1, points along the same direction are not labelled identically, whereas points along \emph{Dir} 2 are assigned to the same class, i.e., satisfying \emph{directionality}. Note that in this study, we say a model can fit a specific point if its prediction logit function (before \emph{Sigmoid}) is negative for $\bullet$ and positive for \textbf{$+$}. While zero-bias networks have a more limited expressive capacity compared to normal networks, being restricted to linear functions originating from the origin, they can effortlessly fit \emph{Dir} 2, as shown in Figure \ref{fig:logit_dir2}. This is due to the fact that all points in the predicted logit fall below 0. However, in the absence of directionality - \emph{Dir} 1, shown in Figure \ref{fig:logit_dir1}, the points in the predicted logit scatter across the 0 line (as they belong to different classes). A linear function starting from the origin could never fit this case. On the other hand, normal networks, with their highly expressive piece-wise functions, can fit both \emph{Dir} 1 and \emph{Dir} 2.

In conclusion, our observations indicate that both types of neural networks achieve similar accuracies in image classification tasks (because of directionality), which is consistent with our experimental findings in Section \ref{sec:scalar_inv_eval}. Moreover, we propose that directionality can serve as a powerful geometric prior in image classification, akin to the translational invariance prior employed in CNNs.

\section{Related work}
\label{sec:related}

\subsection{Invariance in neural networks}

Studying invariance in machine learning as well as neural networks has attracted much attention as real-world data such as images often exhibit rich invariant structures. Incorporating such invariance properties as prior knowledge (inductive bias)  could expand the expressive power of the network without much increase in the number of parameters, which usually leads to better performance. For instance, Convolutional Neural Networks have a stronger geometric prior - translation invariance \cite{geo_dl,cogprints5869}. In addition, Group equivariant Convolutional Neural Networks (G-CNNs) adapt group convolution layers to achieve great results on images generated by translations, reflections, and rotations \cite{pmlr-v48-cohenc16}. Similar work also focuses on studying the invariance of neural network’s outputs under group actions on its inputs \cite{DBLP:conf/icml/KondorT18,Benefits_of_Invariance,DBLP:journals/jmlr/Bloem-ReddyT20}. 


Given the scale invariant nature of images, there is also a line of work studies how to improve the consistency of models' prediction on varying scale images \cite{DBLP:journals/corr/XuXZYZ14,DBLP:journals/corr/abs-1906-03861,invar_,zhang2017s3fd}. However, the most related invariance to our work is illumination invariance which has a great impact on many real-world applications. For example, Ramaiah et al. \cite{7091490} uses convolutional neural networks for face recognition under non-uniform illumination. Maddern et al. \cite{Maddern2014IlluminationII} studies illumination
invariant transform to improve visual localization, mapping, and scene classification for autonomous road vehicles. Huang et al. \cite{DBLP:conf/mm/HuangZF019}
leverages Retinex Decomposition Net and bottom-up attention to approach person re-identification. Despite absolute invariance being considered hard to achieve and most works usually failing to guarantee it, our work shows that \emph{absolute invariance} under scalar multiplication can be achieved with zero-bias neural networks.

\subsection{Zero-bias neural networks}

Although zero-bias neural networks do not appear as much as normal neural networks in the machine-learning literature due to potential reductions in models' expressive capability, they have been used in some real-world applications such as facial expression recognition\cite{khorrami2015deep}, abnormal event detection in IoT\cite{liu2021zero}, identification of Internet-of-Things devices\cite{9173537},  RF signal surveillance\cite{liu2}, and anomaly data detection\cite{bo}. There are several reasons for choosing zero-bias neural networks over normal neural networks: (1) Their incremental learning fashion and better decision fairness; (2) Better interpretability without losing accuracy, which challenges the common first impression of the weaker expressive capability of zero-bias models; (3) More reliable and robust performance. Although these works achieve some success with zero-bias neural networks, none of them dive deeper to analyze these advantages formally. Our work explores zero-bias from an invariant perspective
for the first time, to our best knowledge, 
identifying scalar multiplication invariance in zero-bias models, proving some rigorous robust guarantees, and explaining their comparable accuracy based on geometric sights of image distribution.

\section{Conclusion}
\label{sec:conclusion}


In this paper, we study how neural networks behave after removing bias terms, with a focus on image classification. We prove that, by simply dropping bias terms, the prediction of neural networks achieves absolute invariance under varying contrast of the input image. Moreover, derived from the scalar invariance property, we show that zero-bias networks are robust on certain lines and convex regions of the input space. Although it is commonly believed that bias improves models' expressive capability and thus is always needed, we show that it can be completely ignored for image classification tasks if we consider directionality, an important property of the intrinsic distribution of images. We further show that zero-bias networks and normal networks share almost identical training dynamics through empirical evaluation as well as a theoretical tool called the Neural Tangent Kernel. We also demonstrate the fairness of zero-bias networks' prediction on the zero image - a uniform belief in all candidates. This may provide an orthogonal approach to improving fairness in current AI systems. Finally, we believe dropping bias terms shares the spirit of adapting convolutions as a strong prior in designing neural network architecture in computer vision. We consider this work as a preliminary step towards comprehending zero-bias networks and intend to further explore their robustness, fairness, and generalization ability in our future research.

\clearpage
{
\small


\bibliographystyle{ieee_fullname}
}

\bibliography{main}

\begin{thebibliography}{10}\itemsep=-1pt

\bibitem{ThreatAd}
Naveed Akhtar and Ajmal~S. Mian.
\newblock Threat of adversarial attacks on deep learning in computer vision:
  {A} survey.
\newblock {\em {IEEE} Access}, 6:14410--14430, 2018.

\bibitem{Alzubaidi2021ReviewOD}
Laith Alzubaidi, Jinglan Zhang, Amjad~J. Humaidi, Ayad Al-dujaili, Ye Duan,
  Omran Al-Shamma, Jesus Santamar{\'i}a, Mohammed~Abdulraheem Fadhel, Muthana
  Al-Amidie, and Laith Farhan.
\newblock Review of deep learning: concepts, cnn architectures, challenges,
  applications, future directions.
\newblock {\em Journal of Big Data}, 8, 2021.

\bibitem{DBLP:journals/jmlr/Bloem-ReddyT20}
Benjamin Bloem{-}Reddy and Yee~Whye Teh.
\newblock Probabilistic symmetries and invariant neural networks.
\newblock {\em J. Mach. Learn. Res.}, 21:90:1--90:61, 2020.

\bibitem{cogprints5869}
Jake Bouvrie.
\newblock Notes on convolutional neural networks.
\newblock {\em CoRR}, 2006.

\bibitem{BrockDS21}
Andrew Brock, Soham De, and Samuel~L. Smith.
\newblock Characterizing signal propagation to close the performance gap in
  unnormalized resnets.
\newblock In {\em {ICLR}}. OpenReview.net, 2021.

\bibitem{BrockDSS21}
Andy Brock, Soham De, Samuel~L. Smith, and Karen Simonyan.
\newblock High-performance large-scale image recognition without normalization.
\newblock In {\em {ICML}}, volume 139 of {\em Proceedings of Machine Learning
  Research}, pages 1059--1071. {PMLR}, 2021.

\bibitem{geo_dl}
Michael~M. Bronstein, Joan Bruna, Taco Cohen, and Petar Velickovic.
\newblock Geometric deep learning: Grids, groups, graphs, geodesics, and
  gauges.
\newblock {\em CoRR}, abs/2104.13478, 2021.

\bibitem{buolamwini_2019}
Joy Buolamwini.
\newblock Artificial intelligence has a racial and gender bias problem, Feb
  2019.

\bibitem{nonrobust}
Nicholas Carlini, Anish Athalye, Nicolas Papernot, Wieland Brendel, Jonas
  Rauber, Dimitris Tsipras, Ian~J. Goodfellow, Aleksander Madry, and Alexey
  Kurakin.
\newblock On evaluating adversarial robustness.
\newblock {\em CoRR}, abs/1902.06705, 2019.

\bibitem{pmlr-v48-cohenc16}
Taco Cohen and Max Welling.
\newblock Group equivariant convolutional networks.
\newblock In Maria~Florina Balcan and Kilian~Q. Weinberger, editors, {\em
  Proceedings of The 33rd International Conference on Machine Learning},
  volume~48 of {\em Proceedings of Machine Learning Research}, pages
  2990--2999, New York, New York, USA, 20--22 Jun 2016. PMLR.

\bibitem{imagenet}
Jia Deng, Wei Dong, Richard Socher, Li-Jia Li, Kai Li, and Li Fei-Fei.
\newblock Imagenet: A large-scale hierarchical image database.
\newblock In {\em 2009 IEEE Conference on Computer Vision and Pattern
  Recognition}, pages 248--255, 2009.

\bibitem{MPTs_wh}
James Diffenderfer, Brian~R. Bartoldson, Shreya Chaganti, Jize Zhang, and
  Bhavya Kailkhura.
\newblock A winning hand: Compressing deep networks can improve
  out-of-distribution robustness.
\newblock {\em CoRR}, abs/2106.09129, 2021.

\bibitem{MPTs}
James Diffenderfer and Bhavya Kailkhura.
\newblock Multi-prize lottery ticket hypothesis: Finding accurate binary neural
  networks by pruning {A} randomly weighted network.
\newblock {\em CoRR}, abs/2103.09377, 2021.

\bibitem{ViT}
Alexey Dosovitskiy, Lucas Beyer, Alexander Kolesnikov, Dirk Weissenborn,
  Xiaohua Zhai, Thomas Unterthiner, Mostafa Dehghani, Matthias Minderer, Georg
  Heigold, Sylvain Gelly, Jakob Uszkoreit, and Neil Houlsby.
\newblock An image is worth 16x16 words: Transformers for image recognition at
  scale.
\newblock {\em CoRR}, abs/2010.11929, 2020.

\bibitem{geng2022}
Chuqin Geng, Nham Le, Xiaojie Xu, Zhaoyue Wang, Arie Gurfinkel, and Xujie Si.
\newblock Towards reliable neural specifications, 2022.

\bibitem{DBLP:journals/corr/abs-1906-03861}
Rohan Ghosh and Anupam~K. Gupta.
\newblock Scale steerable filters for locally scale-invariant convolutional
  neural networks.
\newblock {\em CoRR}, abs/1906.03861, 2019.

\bibitem{LR1}
Boris Hanin and David Rolnick.
\newblock Complexity of linear regions in deep networks.
\newblock In {\em {ICML}}, volume~97 of {\em Proceedings of Machine Learning
  Research}, pages 2596--2604. {PMLR}, 2019.

\bibitem{LR2}
Boris Hanin and David Rolnick.
\newblock Deep relu networks have surprisingly few activation patterns.
\newblock In {\em NeurIPS}, pages 359--368, 2019.

\bibitem{HASSABIS2017245}
Demis Hassabis, Dharshan Kumaran, Christopher Summerfield, and Matthew
  Botvinick.
\newblock Neuroscience-inspired artificial intelligence.
\newblock {\em Neuron}, 95(2):245--258, 2017.

\bibitem{RESNET}
Kaiming He, Xiangyu Zhang, Shaoqing Ren, and Jian Sun.
\newblock Deep residual learning for image recognition, 2015.

\bibitem{DBLP:conf/iccv/HeZRS15}
Kaiming He, Xiangyu Zhang, Shaoqing Ren, and Jian Sun.
\newblock Delving deep into rectifiers: Surpassing human-level performance on
  imagenet classification.
\newblock In {\em {ICCV}}, pages 1026--1034. {IEEE} Computer Society, 2015.

\bibitem{xdnn}
Robin Hesse, Simone Schaub{-}Meyer, and Stefan Roth.
\newblock Fast axiomatic attribution for neural networks.
\newblock {\em CoRR}, abs/2111.07668, 2021.

\bibitem{Howard_Imagenette_2019}
Jeremy Howard.
\newblock Imagenette: A smaller subset of 10 easily classified classes from
  imagenet, March 2019.

\bibitem{DBLP:conf/mm/HuangZF019}
Yukun Huang, Zheng{-}Jun Zha, Xueyang Fu, and Wei Zhang.
\newblock Illumination-invariant person re-identification.
\newblock In {\em {ACM} Multimedia}, pages 365--373. {ACM}, 2019.

\bibitem{Batchnorm}
Sergey Ioffe and Christian Szegedy.
\newblock Batch normalization: Accelerating deep network training by reducing
  internal covariate shift.
\newblock In {\em {ICML}}, volume~37 of {\em {JMLR} Workshop and Conference
  Proceedings}, pages 448--456. JMLR.org, 2015.

\bibitem{ntk2018}
Arthur Jacot, Franck Gabriel, and Cl{\'{e}}ment Hongler.
\newblock Neural tangent kernel: Convergence and generalization in neural
  networks.
\newblock {\em CoRR}, abs/1806.07572, 2018.

\bibitem{bio_neuro}
A.K. Jain, Jianchang Mao, and K.M. Mohiuddin.
\newblock Artificial neural networks: a tutorial.
\newblock {\em Computer}, 29(3):31--44, 1996.

\bibitem{khorrami2015deep}
Pooya Khorrami, Thomas Paine, and Thomas Huang.
\newblock Do deep neural networks learn facial action units when doing
  expression recognition?
\newblock In {\em Proceedings of the IEEE international conference on computer
  vision workshops}, pages 19--27, 2015.

\bibitem{DBLP:conf/icml/KondorT18}
Risi Kondor and Shubhendu Trivedi.
\newblock On the generalization of equivariance and convolution in neural
  networks to the action of compact groups.
\newblock In {\em {ICML}}, volume~80 of {\em Proceedings of Machine Learning
  Research}, pages 2752--2760. {PMLR}, 2018.

\bibitem{leavy2018gender}
Susan Leavy.
\newblock Gender bias in artificial intelligence: The need for diversity and
  gender theory in machine learning.
\newblock In {\em Proceedings of the 1st international workshop on gender
  equality in software engineering}, pages 14--16, 2018.

\bibitem{Liu2017CVPR}
Weiyang Liu, Yandong Wen, Zhiding Yu, Ming Li, Bhiksha Raj, and Le Song.
\newblock Sphereface: Deep hypersphere embedding for face recognition.
\newblock In {\em CVPR}, 2017.

\bibitem{Liu2017NIPS}
Weiyang Liu, Yan-Ming Zhang, Xingguo Li, Zhiding Yu, Bo Dai, Tuo Zhao, and Le
  Song.
\newblock Deep hyperspherical learning.
\newblock In {\em NIPS}, 2017.

\bibitem{liu2}
Yongxin Liu, Yingjie Chen, Jian Wang, Shuteng Niu, Dahai Liu, and Houbing Song.
\newblock Zero-bias deep neural network for quickest rf signal surveillance.
\newblock In {\em 2021 IEEE International Performance, Computing, and
  Communications Conference (IPCCC)}, pages 1--8, 2021.

\bibitem{liu2021zero}
Yongxin Liu, Jian Wang, Jianqiang Li, Shuteng Niu, Lei Wu, and Houbing Song.
\newblock Zero-bias deep learning enabled quickest abnormal event detection in
  iot.
\newblock {\em IEEE Internet of Things Journal}, 2021.

\bibitem{9173537}
Yongxin Liu, Jian Wang, Jianqiang Li, Houbing Song, Thomas Yang, Shuteng Niu,
  and Zhong Ming.
\newblock Zero-bias deep learning for accurate identification of
  internet-of-things (iot) devices.
\newblock {\em IEEE Internet of Things Journal}, 8(4):2627--2634, 2021.

\bibitem{Benefits_of_Invariance}
Clare Lyle, Mark van~der Wilk, Marta Kwiatkowska, Yarin Gal, and Benjamin
  Bloem{-}Reddy.
\newblock On the benefits of invariance in neural networks.
\newblock {\em CoRR}, abs/2005.00178, 2020.

\bibitem{Maddern2014IlluminationII}
Will Maddern, Alex Stewart, Colin McManus, Ben Upcroft, Winston Churchill, and
  Paul Newman.
\newblock Illumination invariant imaging: Applications in robust vision-based
  localisation, mapping and classification for autonomous vehicles.
\newblock In {\em Proceedings of the Visual Place Recognition in Changing
  Environments Workshop, IEEE International Conference on Robotics and
  Automation (ICRA), Hong Kong, China}, volume~2, page~5, 2014.

\bibitem{mehrabi2021survey}
Ninareh Mehrabi, Fred Morstatter, Nripsuta Saxena, Kristina Lerman, and Aram
  Galstyan.
\newblock A survey on bias and fairness in machine learning.
\newblock {\em ACM Computing Surveys (CSUR)}, 54(6):1--35, 2021.

\bibitem{MONTAVON20181}
Grégoire Montavon, Wojciech Samek, and Klaus-Robert Müller.
\newblock Methods for interpreting and understanding deep neural networks.
\newblock {\em Digital Signal Processing}, 73:1--15, 2018.

\bibitem{invar_}
S.J. Perantonis and P.J.G. Lisboa.
\newblock Translation, rotation, and scale invariant pattern recognition by
  high-order neural networks and moment classifiers.
\newblock {\em IEEE Transactions on Neural Networks}, 3(2):241--251, 1992.

\bibitem{7091490}
N.~Pattabhi Ramaiah, Earnest~Paul Ijjina, and C.~Krishna Mohan.
\newblock Illumination invariant face recognition using convolutional neural
  networks.
\newblock In {\em 2015 IEEE International Conference on Signal Processing,
  Informatics, Communication and Energy Systems (SPICES)}, pages 1--4, 2015.

\bibitem{ShamARKAOOA23}
Abdallah~Hussein Sham, Kadir Aktas, Davit Rizhinashvili, Danila Kuklianov,
  Fatih Alisinanoglu, Ikechukwu Ofodile, Cagri Ozcinar, and Gholamreza
  Anbarjafari.
\newblock Ethical {AI} in facial expression analysis: racial bias.
\newblock {\em Signal Image Video Process.}, 17(2):399--406, 2023.

\bibitem{Dropout}
Nitish Srivastava, Geoffrey~E. Hinton, Alex Krizhevsky, Ilya Sutskever, and
  Ruslan Salakhutdinov.
\newblock Dropout: a simple way to prevent neural networks from overfitting.
\newblock {\em J. Mach. Learn. Res.}, 15(1):1929--1958, 2014.

\bibitem{EfficientNet}
Mingxing Tan and Quoc~V. Le.
\newblock Efficientnet: Rethinking model scaling for convolutional neural
  networks.
\newblock {\em CoRR}, abs/1905.11946, 2019.

\bibitem{instancenormalization}
Dmitry Ulyanov, Andrea Vedaldi, and Victor~S. Lempitsky.
\newblock Instance normalization: The missing ingredient for fast stylization.
\newblock {\em CoRR}, abs/1607.08022, 2016.

\bibitem{wang2019bias}
Shengjie Wang, Tianyi Zhou, and Jeff Bilmes.
\newblock Bias also matters: Bias attribution for deep neural network
  explanation.
\newblock In {\em International Conference on Machine Learning}, pages
  6659--6667. PMLR, 2019.

\bibitem{weng2022ntk}
Lilian Weng.
\newblock Some math behind neural tangent kernel.
\newblock {\em Lil'Log}, Sep 2022.

\bibitem{xu2020neural}
Keyulu Xu, Mozhi Zhang, Jingling Li, Simon~Shaolei Du, Ken{-}ichi
  Kawarabayashi, and Stefanie Jegelka.
\newblock How neural networks extrapolate: From feedforward to graph neural
  networks.
\newblock In {\em 9th International Conference on Learning Representations,
  {ICLR} 2021, Virtual Event, Austria, May 3-7, 2021}. OpenReview.net, 2021.

\bibitem{DBLP:journals/corr/XuXZYZ14}
Yichong Xu, Tianjun Xiao, Jiaxing Zhang, Kuiyuan Yang, and Zheng Zhang.
\newblock Scale-invariant convolutional neural networks.
\newblock {\em CoRR}, abs/1411.6369, 2014.

\bibitem{YANG20201048}
Guangyu~Robert Yang and Xiao-Jing Wang.
\newblock Artificial neural networks for neuroscientists: A primer.
\newblock {\em Neuron}, 107(6):1048--1070, 2020.

\bibitem{bo}
Bo Zhang, Qiang Zhang, Yong~Xin Liu, and Ou Ye.
\newblock Anomaly data detection for ads- b based on zero-bias inception
  network.
\newblock In {\em 2021 IEEE International Conference on Signal Processing,
  Communications and Computing (ICSPCC)}, pages 1--6, 2021.

\bibitem{mixup_interpolation}
Hongyi Zhang, Moustapha Ciss{\'{e}}, Yann~N. Dauphin, and David Lopez{-}Paz.
\newblock mixup: Beyond empirical risk minimization.
\newblock {\em CoRR}, abs/1710.09412, 2017.

\bibitem{ZhangDM19}
Hongyi Zhang, Yann~N. Dauphin, and Tengyu Ma.
\newblock Fixup initialization: Residual learning without normalization.
\newblock In {\em {ICLR} (Poster)}. OpenReview.net, 2019.

\bibitem{zhang2017s3fd}
Shifeng Zhang, Xiangyu Zhu, Zhen Lei, Hailin Shi, Xiaobo Wang, and Stan~Z Li.
\newblock S3fd: Single shot scale-invariant face detector.
\newblock In {\em Proceedings of the IEEE international conference on computer
  vision}, pages 192--201, 2017.

\end{thebibliography}

\newpage
\onecolumn
\appendix

\section{Proof of interpolation robustness property}
\label{sec: proof_interpolation_robustness_property}


\textbf{Theorem} \ref{lemma:interpolation_robustness_property}\textbf{(Interpolation robustness property)}
\emph{For any two inputs $X_1$ and $X_2$ that have the same prediction and neural activation pattern by network $\mathcal{N}$, i.e., $\mathcal{N}(X_1) = \mathcal{N}(X_2)$ and $\cP_{X_1} = \cP_{X_2}$, their linear interpolation also yield the same prediction, that is, $\mathcal{N}(\lambda X_1 + (1-\lambda)X_2) = \mathcal{N}(X_1) = \mathcal{N}(X_2)$, where $\lambda \in [0,1]$.}

\textbf{Proof:}
We show the interpolation robustness property holds for fully connected neural networks without bias. For more complicated neural networks such as CNN, the property also holds as long as all transformations before the output layer are scalar associative (\textbf{Lemma} \ref{lemma: directional_robustness_property}).
Consider a FCN $\mathcal{N}$ composed of $J$ number of fully connected layers $\mathcal{L}_j$ and some \textit{ReLU} layers $\mathcal{R}$. We think of layers and activation functions as transformations on the input $X$, then the output of the network $\mathcal{O}(\lambda X_1 + (1-\lambda)X_2)$ before the \textit{softamx} function is represented by:
\begin{equation}
\label{eqn:output_before_softmax}
\mathcal{O}(\lambda X_1 + (1-\lambda)X_2)  = \mathcal{L}_J \circ \mathcal{R} \circ ... \circ \mathcal{R} \circ \mathcal{L}_1 \circ (\lambda X_1 + (1-\lambda)X_2) 
\end{equation}

For any fully connected layer $\mathcal{L}_j$, we have:
\begin{equation}
\label{eqn:output_for_FLC_layer}
\mathcal{L}_j \circ (\lambda X_1 + (1-\lambda)X_2) =  (\lambda X_1 + (1-\lambda)X_2)\mathcal{W}_j^{T} = \lambda X{W}_j^{T} + (1-\lambda)X_2\mathcal{W}_j^{T} = \lambda  \mathcal{L}_j \circ X_1 +  (1-\lambda) \mathcal{L}_j \circ X_2
\end{equation}

On the other hand, we have $X$ and $Y$ falling into the same neural activation pattern. Since the linear region corresponding to the neural activation pattern is convex, the interpolation of $X_1$ and $X_2$, $\lambda X_1 + (1-\lambda)X_2$, also lies in the same neural activation pattern. Furthermore, we have:
\begin{equation}
\begin{gathered}
\mathcal{R} \circ \mathcal{L}_1 \circ (\lambda X_1 + (1-\lambda)X_2) =  \lambda  \mathcal{R} \circ \mathcal{L}_1 \circ X_1 +  (1-\lambda) \mathcal{R} \circ \mathcal{L}_1 \circ X_2\\
\mathcal{R} \circ \mathcal{L}_2 \circ (\lambda  \mathcal{R} \circ \mathcal{L}_1 \circ X_1 +  (1-\lambda) \mathcal{R} \circ \mathcal{L}_1 \circ X_2 ) =  \lambda  \mathcal{R} \circ \mathcal{L}_2 \circ  \mathcal{R} \circ \mathcal{L}_1 \circ X +  (1-\lambda) \mathcal{R} \circ \mathcal{L}_2 \circ \mathcal{R} \circ \mathcal{L}_1 \circ X_2\\
 ... \\
\mathcal{O}(\lambda X_1 + (1-\lambda)X_2)  = \lambda \mathcal{O}(X_1) + (1-\lambda)\mathcal{O}(X_2), \text{by \textbf{Lemma}\ref{lemma: directional_robustness_property}}
\end{gathered}
\end{equation}

Given that $\mathcal{N}(X_1) = \mathcal{N}(X_2)$, the index/class of the highest logit of $\mathcal{O}(X_1)$ and $\mathcal{O}(X_2)$ must be the same, that is:
\begin{equation}
\label{eqn}
\argmax_{c} \mathcal{O}(X_1)_{c} = \argmax_{c} \mathcal{O}(X_2)_{c}
\end{equation}

Since multiplying a positive scalar to the operand won't change the output of the$\argmax$ operator, we have:
\begin{equation}
\label{eqn:interpolation_robustness_property_eqn}
\argmax_{c} \lambda\mathcal{O}(X_1)_{c} = \argmax_{c} (1-\lambda)\mathcal{O}(X_2)_{c}= \argmax_{c} \mathcal{O}(X_1)_{c} = \argmax_{c} \mathcal{O}(X_2)_{c}
\end{equation}

Note that the index/class of the highest logit of $ \lambda\mathcal{O}(X_1)_{c}$ and $(1-\lambda)\mathcal{O}(X_2)_{c}$ are the same, the index/class of the highest logit of their addition is also the same as $ \lambda\mathcal{O}(X_1)_{c}$ and $(1-\lambda)\mathcal{O}(X_2)_{c}$. Then it follows that:
\begin{equation}    
\argmax_{c} \mathcal{O}(\lambda X_1 + (1-\lambda)X_2)_{c}= \argmax_{c} \mathcal{O}(X_1)_{c} = \argmax_{c} \mathcal{O}(X_2)_{c}
\end{equation}

Since the \textit{softmax} function will preserve the ranking of logits, we have:
\begin{equation}
\label{eqn:softmax_preserve_rank}
\argmax_{c} \frac{e^{\mathcal{O}(\lambda X_1 + (1-\lambda)X_2)_{c}}}{\displaystyle\sum_{c\in\mathcal{C}} e^{\mathcal{O}(\lambda X_1 + (1-\lambda)X_2)_{c}}} 
 = \argmax_{c} \frac{e^{\mathcal{O}(X_1)_{c}}}{\displaystyle\sum_{c \in \mathcal{C}} e^{\mathcal{O}(X_1)_{c}}} =  \argmax_{c} \frac{e^{\mathcal{O}(X_2)_{c}}}{\displaystyle\sum_{c \in \mathcal{C}} e^{\mathcal{O}(X_2)_{c}}} 
\end{equation}

Finally, this can be restated as:
\begin{equation}
\label{eqn:Interpolation_robustness_property_conclusion}
\mathcal{N}(\lambda X + (1-\lambda)X_2) = \mathcal{N}(X_1) = \mathcal{N}(X_2)
\end{equation}



\newpage

\section{Batch normalization free methods }
\label{sec: batch_normal}
\textbf{Fixup} enables training
deep residual networks with comparable performance in terms of convergence, generalization, etc, without normalization. More specifically, this method rescales the standard initialization of residual branches by taking the network architecture into account. The key steps of Fixup initialization are described as follows:
\begin{enumerate}
    \item Initialize the last layer of each residual branch and the classification layer to 0.
    \item Initialize other layers using a standard method \cite{DBLP:conf/iccv/HeZRS15}, and scale only the weight layers inside residual branches by $L^{-\frac{1}{2m-2}}$, where $L$ and $m$ are the numbers of residual blocks and layers inside a residual branch respectively.
    \item Add a scalar multiplier before each convolution, linear, and element-wise activation layer in each residual branch, the multiplier is initialized at $1$ \footnote{We intentionally ignore the scalar bias (initialized at 0) presented in the original paper to ensure scalar invariance. }.
\end{enumerate}
It is obvious that the above initialization steps perform some transformations on the weights of neural networks instead of the input, and the scalar multiplier is scalar associative which ensures the trained ResNet is scalar invariant.

\textbf{NFNet} aims to overcome the same challenge of developing ResNet variants without normalization layers yet is comparable to batch-normalized ResNets in many aspects. The effect of standard batch normalization operation within each residual block can be summarized as: 1) downscales the input by a factor proportional to its standard deviation; 2)  increases the variance of the input signal by an approximately constant factor. By mimicking the effect of batch normalization, the residual blocks can be written in the form of $X_{l+1} = X_l + \alpha \mathcal{G}_l( X_l / \beta_l)$,  where $X_l$ denotes the input to the $l^{th}$ residual block and $\mathcal{G}_l(\cdot)$ denotes the $l^{th}$ residual branch. Moreover, the network should be designed such that:

\begin{itemize}
    \item $\mathcal{G}_l(\cdot)$ is parameterized to be able to preserve variance at initialization, i.e., $Var(\mathcal{G}_l(z)) = Var(z)$ for all $l$.
    \item $\beta_l$ is a fixed scalar, set it to be $\sqrt{Var(X_l)}$, the expected empirical standard deviation of $X_l$ at initialization. 
    \item $\alpha$ is a hyperparameter that controls the growth rate of variance between blocks.
\end{itemize}
Since both $\alpha$ and $\beta$ are fixed scalar during the inference phase. The modified residual blocks are  scalar associative since $s X_l + \alpha \mathcal{G}_l( s X_l / \beta_l) = s (X_l + \alpha \mathcal{G}_l( X_l / \beta_l))$. We conclude the NFNet method also ensures scalar invariance.

\section{Training details}
\label{sec: training_details}
Our ResNet50 model, as shown in Table \ref{tab:scalar_inv_tab}, is trained for 100 epochs on the training split of ImageNet \cite{Howard_Imagenette_2019} using a NVIDIA A100 (40GB) GPU. We follow the same cosine annealing learning rate scheduler from \cite{xdnn} with an initial learning rate of 0.1 and use a mixup interpolation proposed in \cite{mixup_interpolation} with an interpolation strength $\alpha$ = 0.7 and a batch size of 256, where each epoch takes approximately 40 minutes to complete. Our model is trained with SGD using a 0.9 momentum, and a 1e-4 weight decay. To remove the bias, we mainly employ the Fixup approach and utilize code from two sources \footnote{\href{https://github.com/visinf/fast-axiomatic-attribution}{https://github.com/visinf/fast-axiomatic-attribution}}\footnote{\href{https://github.com/hongyi-zhang/Fixup}{https://github.com/hongyi-zhang/Fixup}}, which are released under the Apache 2.0 and BSD 3-Clause licenses, respectively. In the CIFAR-100 experiment, we utilize a variant of ResNet18 in which we intentionally omitted the Batch Normalization layers. We train the models with a batch size of 128 for 40 epochs, using the same other hyperparameters.

In accordance with the information provided in Table \ref{tab:scalar_inv_tab}, our FCN and CNN models are trained using the following configurations: The FCN model undergoes 20 epochs of training on the training split of the MNIST dataset, utilizing a learning rate of 0.01 and a batch size of 512. On the other hand, the CNN model is trained for 20 epochs on the training split of the Fashion-MNIST dataset, employing a learning rate of 0.01 and a batch size of 128. Both models are trained with Adadelta optimizer.



\newpage

\section{Interpolation robustness examples}
\label{sec: interpolation_robustness_examples}
\begin{figure}[H]
     \centering
         \includegraphics[width=\textwidth]{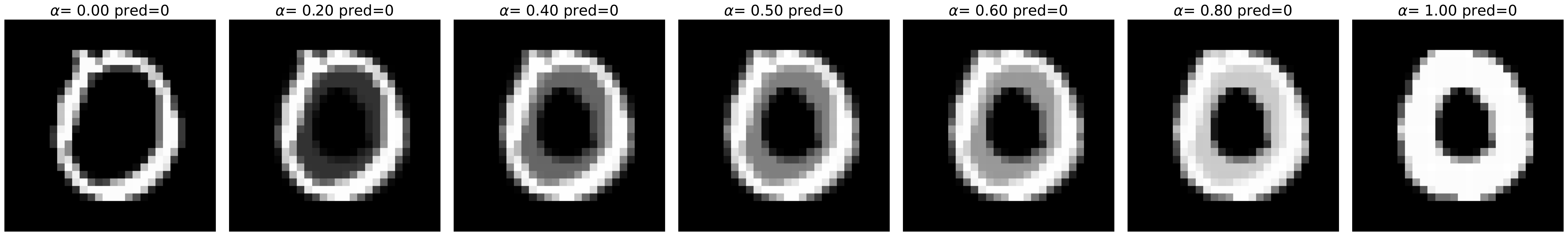}
         \label{fig:Interp_mnist_good_apx_ex1}
         \centering
         \includegraphics[width=\textwidth]{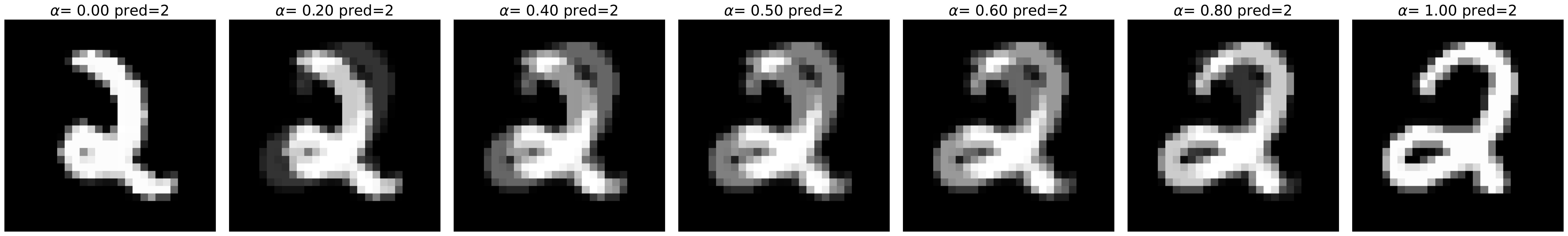}
         \label{fig:Interp_mnist_good_apx_ex2} 
         \centering
         \includegraphics[width=\textwidth]{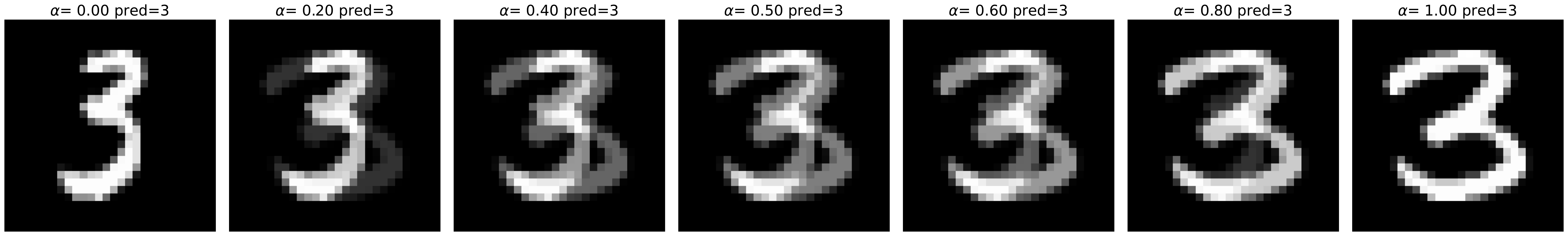}
         \label{fig:Interp_mnist_good_apx_ex3} 
         \centering
         \includegraphics[width=\textwidth]{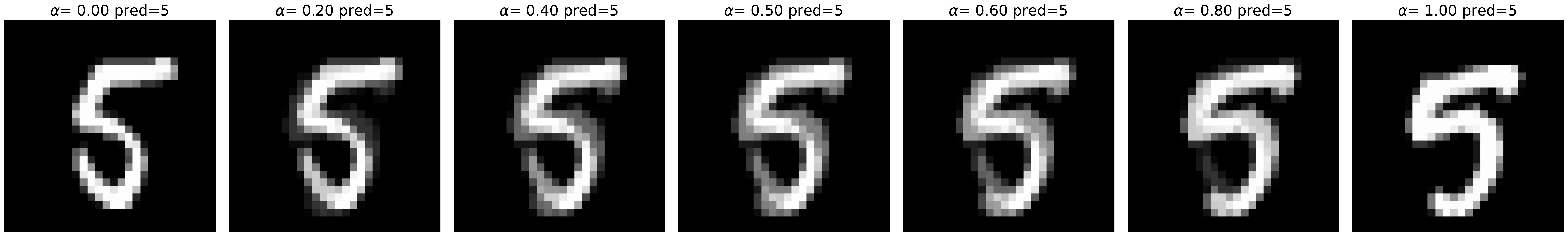}
         \label{fig:Interp_mnist_good_apx_ex4} 
         \centering
         \includegraphics[width=\textwidth]{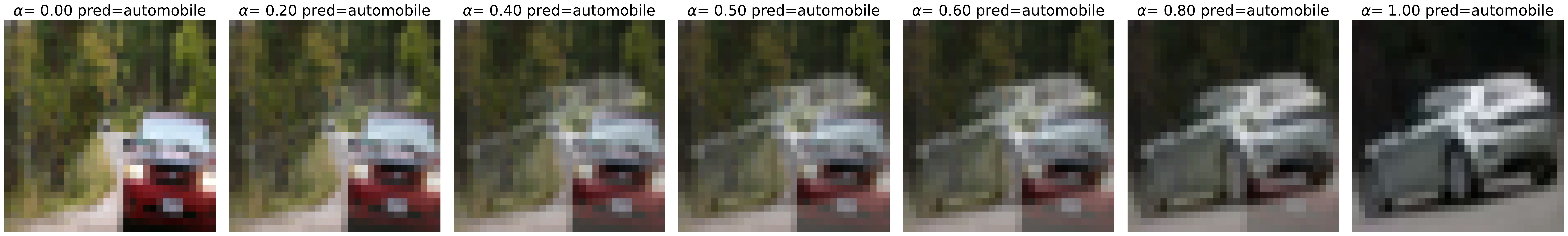}
         \label{fig:Interp_cifar10_good_apx_ex1}
         \centering
         \includegraphics[width=\textwidth]{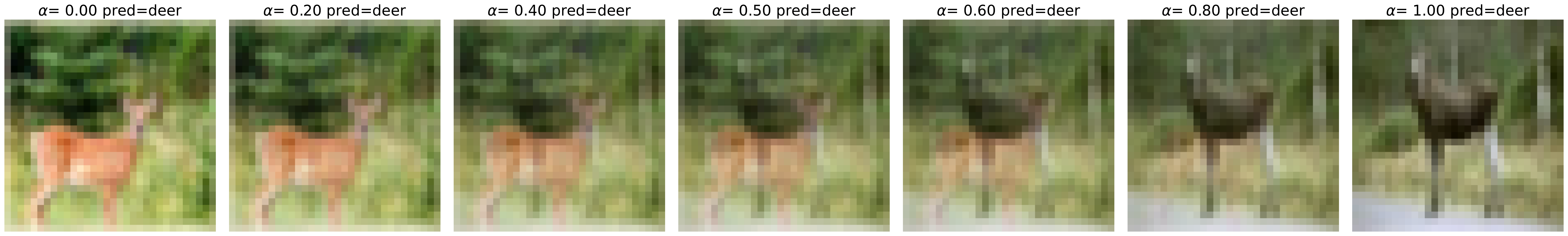}
         \label{fig:Interp_cifar10_good_apx_ex2} 
         \centering
         \includegraphics[width=\textwidth]{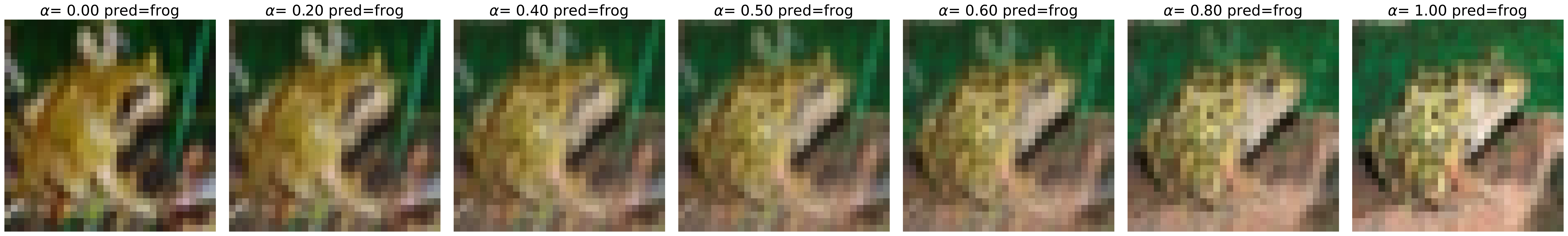}
         \label{fig:Interp_cifar10_good_apx_ex3} 
         \centering
         \includegraphics[width=\textwidth]{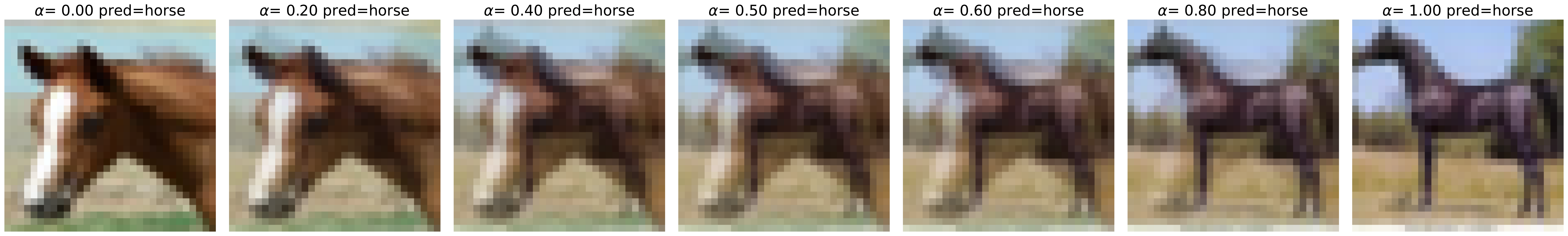}
         \label{fig:Interp_cifar10_good_apx_ex4} 
        \caption{ The left-most and right-most images are from the original MNIST (first five rows) and CIFAR10 (the last five rows) dataset, whereas synthesized/interpolated images are in the middle. For instance, the middle image in the first row is generated by adding $(\alpha = 0.5)$ times the left image to $(1-\alpha)$ times the right image. The interpolated images have the predictions same as the ground truth.}
        \label{fig:Interpolation_good_apx}
\end{figure}

\begin{figure}[H]
     \centering
         \includegraphics[width=\textwidth]{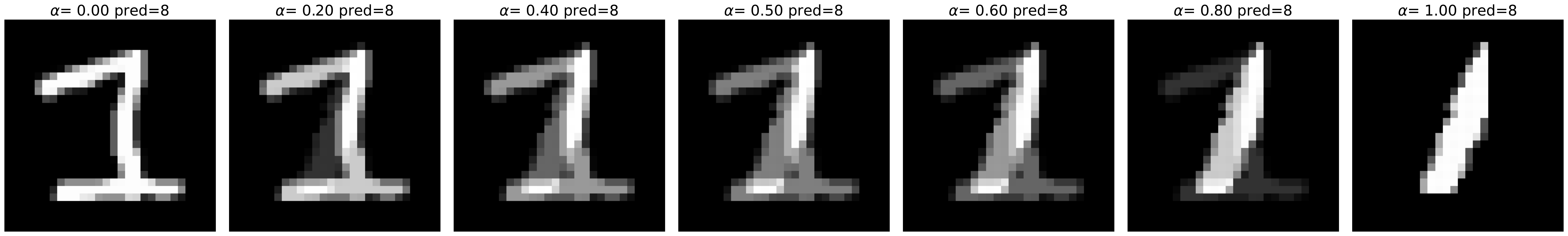}
         \label{fig:Interp_mnist_bad_apx_ex1}
         \centering
         \includegraphics[width=\textwidth]{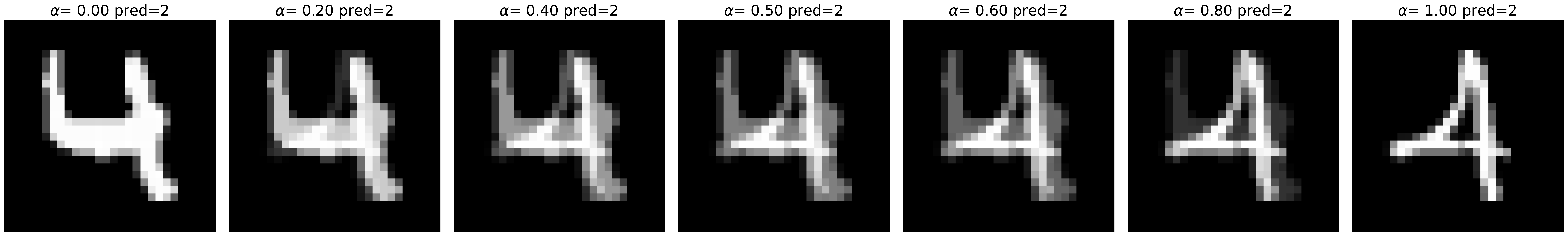}
         \label{fig:Interp_mnist_bad_apx_ex2} 
         \centering
         \includegraphics[width=\textwidth]{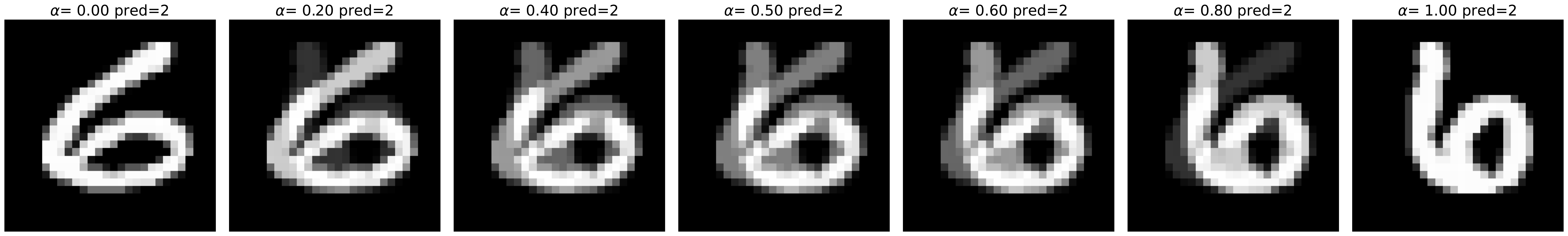}
         \label{fig:Interp_mnist_bad_apx_ex3} 
         \centering
         \includegraphics[width=\textwidth]{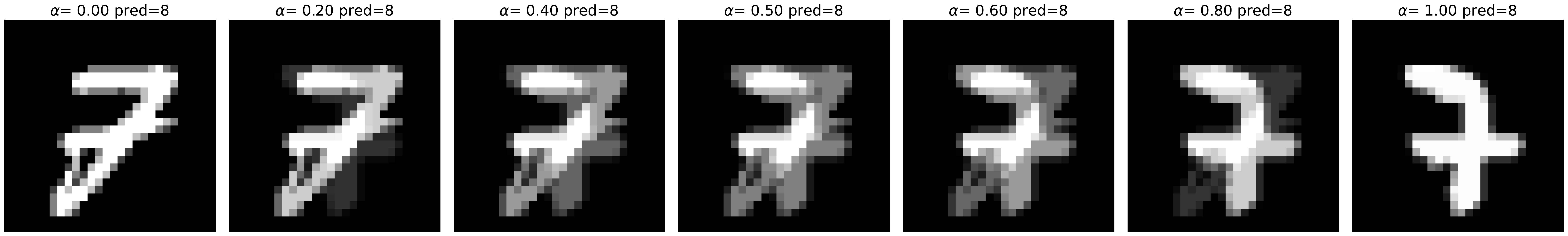}
         \label{fig:Interp_mnist_bad_apx_ex4} 
         \centering
         \includegraphics[width=\textwidth]{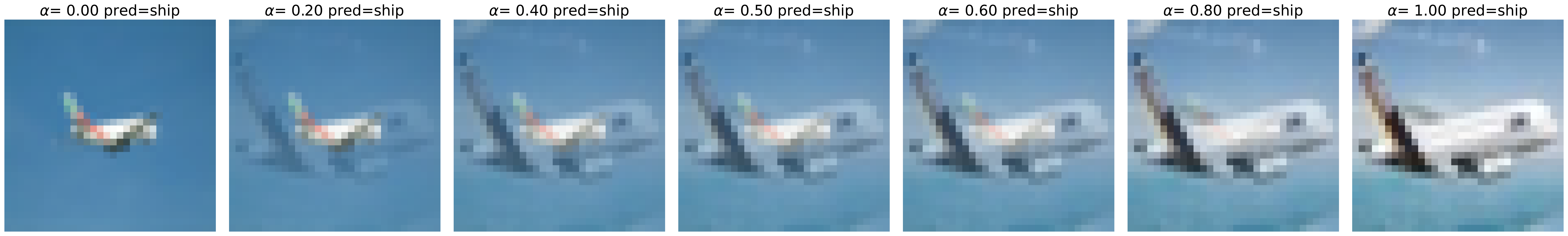}
         \label{fig:Interp_cifar10_bad_apx_ex1}
         \centering
         \includegraphics[width=\textwidth]{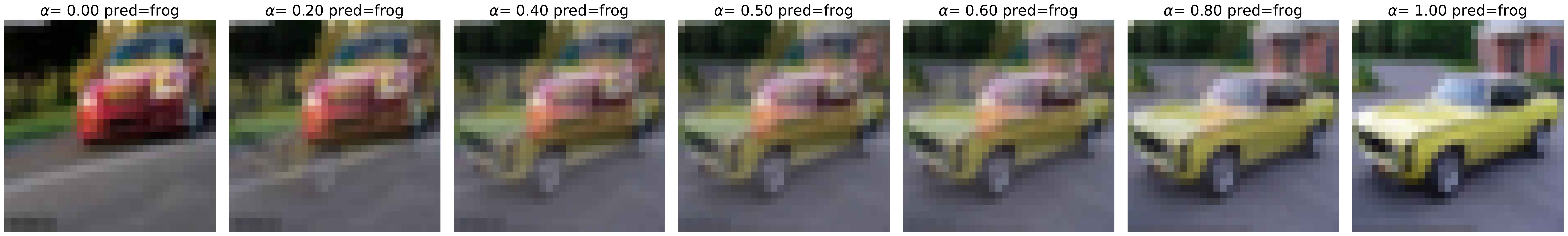}
         \label{fig:Interp_cifar10_bad_apx_ex2} 
         \centering
         \includegraphics[width=\textwidth]{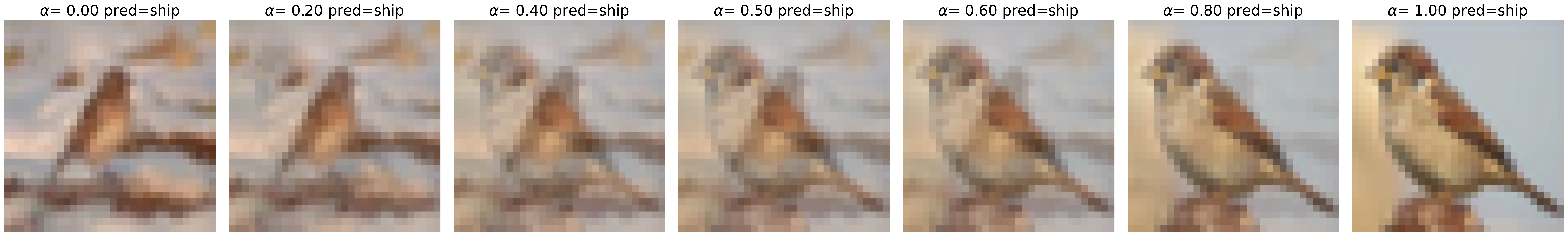}
         \label{fig:Interp_cifar10_bad_apx_ex3} 
         \centering
         \includegraphics[width=\textwidth]{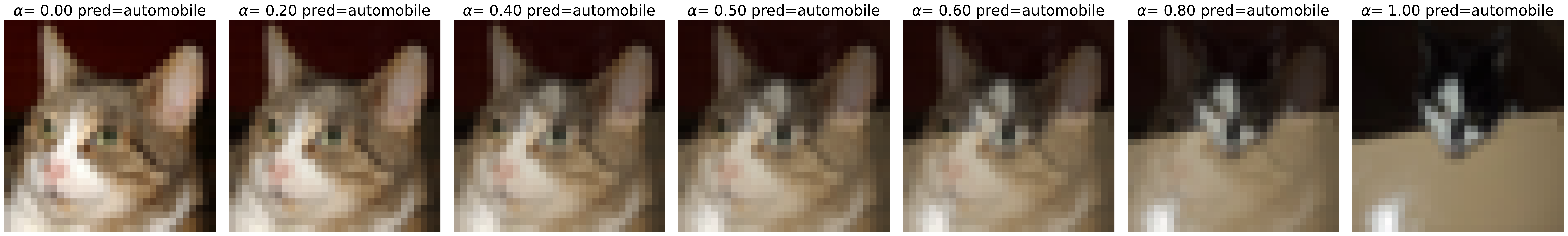}
         \label{fig:Interp_cifar10_bad_apx_ex4} 
        \caption{ The left-most and right-most images are from the original MNIST (first five rows) and CIFAR10 (the last five rows) dataset, whereas synthesized/interpolated images are in the middle. For instance, the middle image in the first row is generated by adding $(\alpha = 0.5)$ times the left image to $(1-\alpha)$ times the right image. The synthesized/interpolated images have the predictions different from the ground truth.}
        \label{fig:Interpolation_bad_apx}
\end{figure}

In this section, we show some interpolation of images from MNIST and CIFAR10 datasets. Based on our experiments, we observe that sharing the same neural activation pattern is a more stringent condition than having the same prediction label. To enforce this constraint, we opt for smaller networks with fewer neurons. However, this can lead to lower accuracy. To provide a visual illustration of the result, we present some examples of correctly predicted images in Figure \ref{fig:Interpolation_good_apx}, and examples of incorrectly predicted images in Figure \ref{fig:Interpolation_bad_apx}.

 \newpage



\newpage
\section{Remarks on training dynamics}
In this section, we demonstrate the analysis of the Neural Tangent Kernel to support our remarks in Section \ref{sec: remarks}. The majority of the formulations and theoretical results are taken from the original paper \cite{ntk2018} and the excellent blog post \cite{weng2022ntk}. We begin by introducing some fundamental notations of neural networks, the Neural Tangent Kernel, and Gaussian Processes. However, readers who are already familiar with these concepts may skip this part and proceed directly to Section \ref{sec: key_results} to view our main results.
\subsection{Notation}

Consider a fully-connected neural network with parameter $\theta$, denoted as $f(.;\theta): \mathbb{R}^{n_0} \to \mathbb{R}^{n_L}$, where the layers are indexed from 0 (input) to $L$ (output). Each layer contains $n_0, \dots, n_L$ neurons, with the input size being $n_0$ and the output size being $n_L$. The total number of parameters in the network is given by $P = \sum_{l=0}^{L-1} (n_l + 1) n_{l+1}$, which means $\theta \in \mathbb{R}^P$.

The training dataset consists of $N$ data points, denoted as $\mathcal{D}={\mathbf{x}^{(i)}, y^{(i)}}{i=1}^N$, where the inputs are represented by $\mathcal{X}={\mathbf{x}^{(i)}}{i=1}^N$ and the labels are represented by $\mathcal{Y}={y^{(i)}}_{i=1}^N$.

For the forward pass computation in each layer, we define an affine transformation $A^{(l)}$ for $l=0, \dots, L-1$, with a weight matrix $\mathbf{w}^{(l)} \in \mathbb{R}^{n_{l} \times n_{l+1}}$, a bias term $\mathbf{b}^{(l)} \in \mathbb{R}^{n_{l+1}}$, and a pointwise nonlinearity function $\sigma(.)$ that is Lipschitz continuous.

\begin{equation}
    \begin{aligned}
A^{(0)} &= \mathbf{x} \\
\tilde{A}^{(l+1)}(\mathbf{x}) &= \frac{1}{\sqrt{n_l}} {\mathbf{w}^{(l)}}^\top A^{(l)} + \beta\mathbf{b}^{(l)}\quad\in\mathbb{R}^{n_{l+1}} & \text{; pre-activations}\\
A^{(l+1)}(\mathbf{x}) &= \sigma(\tilde{A}^{(l+1)}(\mathbf{x}))\quad\in\mathbb{R}^{n_{l+1}} & \text{; post-activations}
\end{aligned}
\end{equation}

Note that in the NTK parameterization, the transformation is rescaled by $1/\sqrt{n_l}$ to avoid divergence with infinite-width networks. And the contribution of the bias terms is controlled by the scalar constant $\beta \geq 0$. \textbf{
Note that for zero-bias neural networks, the bias term $\mathbf{b}^{(l)}$ is simply set to ${0}$. Accordingly, the scalar constant $\beta = 0$ in this case.} All network parameters are initialized as i.i.d Gaussian with mean 0 and standard deviation 1 in the following analysis.

\subsection{Neural Tangent Kernel}
Neural Tangent Kernel (NTK) is a crucial concept in understanding neural network training via gradient descent. Essentially, it explains how updating the model parameters using one data sample impacts the predictions for others during neural network training via gradient descent.

Let's begin by breaking down the intuition behind NTK step by step.

During training, the empirical loss function $\mathcal{L}: \mathbb{R}^P \to \mathbb{R}_+$ is defined as follows, using a per-sample cost function $\ell$: $\mathbb{R}^{n_0} \times \mathbb{R}^{n_L} \to \mathbb{R}_+$: 
\begin{equation}
\mathcal{L}(\theta) =\frac{1}{N} \sum_{i=1}^N \ell(f(\mathbf{x}^{(i)}; \theta), y^{(i)})
\end{equation}
and, applying the chain rule,  the gradient of the loss is:
\begin{equation}
    \nabla_\theta \mathcal{L}(\theta)= \frac{1}{N} \sum_{i=1}^N \underbrace{\nabla_\theta f(\mathbf{x}^{(i)}; \theta)}_{\text{size }P \times n_L} 
\underbrace{\nabla_f \ell(f, y^{(i)})}_{\text{size } n_L \times 1}
\end{equation}
Note that each gradient descent update introduces a small incremental change of infinitesimal step size when we track the evolution of the network parameter $\theta$ over time. Due to the smallness of the update step, it can be approximated as a derivative along the time dimension:
\begin{equation}
    \frac{d\theta}{d t} = - \nabla_\theta\mathcal{L}(\theta)  = -\frac{1}{N} \sum_{i=1}^N \nabla_\theta f(\mathbf{x}^{(i)}; \theta) \nabla_f \ell(f, y^{(i)})
\end{equation}

Once more, applying the chain rule, the output of the network evolves according to the derivative:
\begin{equation}
    \frac{df(\mathbf{x};\theta)}{dt} 
= \frac{df(\mathbf{x};\theta)}{d\theta}\frac{d\theta}{dt}
= -\frac{1}{N} \sum_{i=1}^N \underbrace{\nabla_\theta f(\mathbf{x};\theta)^\top \nabla_\theta f(\mathbf{x}^{(i)}; \theta)}_\text{Neural tangent kernel} \color{black}{\nabla_f \ell(f, y^{(i)})}
\end{equation}
Here we define the Neural Tangent Kernel (NTK) as the blue part in the above formula, $K:
\mathbb{R}^{n_0}\times\mathbb{R}^{n_0} \to \mathbb{R}^{n_L \times n_L}:$
\begin{equation}
    K(\mathbf{x}, \mathbf{x}'; \theta) = \nabla_\theta f(\mathbf{x};\theta)^\top \nabla_\theta f(\mathbf{x}'; \theta)
\end{equation}
Each entry of the output matrix $K$ at position $(m, n)$, where $1 \leq m, n \leq n_L$, is given by:
\begin{equation}
    K_{m,n}(\mathbf{x}, \mathbf{x}'; \theta) = \sum_{p=1}^P \frac{\partial f_m(\mathbf{x};\theta)}{\partial \theta_p} \frac{\partial f_n(\mathbf{x}';\theta)}{\partial \theta_p}
\end{equation}

The feature map of an input $\mathbf{x}$ is given by $\varphi(\mathbf{x}) = \nabla_\theta f(\mathbf{x};\theta)$.

\subsection{\textbf{Infinite Width Networks}}

To understand why the effect of one gradient descent update is similar for different initializations of network parameters, several pioneering theoretical works start with infinite-width networks. We will now examine the detailed proof using NTK to understand how it guarantees that infinite-width networks can converge to a global minimum when trained to minimize an empirical loss.

\subsection{Connection with Gaussian Processes}
Deep neural networks are deeply connected with Gaussian processes. The output functions of an $L$-layer network, $f_i(\mathbf{x}; \theta)$, are i.i.d. centered Gaussian processes with covariance $\Sigma^{(L)}$, defined recursively as:
\begin{equation}
    \begin{aligned}
\Sigma^{(1)}(\mathbf{x}, \mathbf{x}') &= \frac{1}{n_0}\mathbf{x}^\top{\mathbf{x}'} + \beta^2 \\
\lambda^{(l+1)}(\mathbf{x}, \mathbf{x}') &= \begin{bmatrix}
\Sigma^{(l)}(\mathbf{x}, \mathbf{x}) & \Sigma^{(l)}(\mathbf{x}, \mathbf{x}') \\
\Sigma^{(l)}(\mathbf{x}', \mathbf{x}) & \Sigma^{(l)}(\mathbf{x}', \mathbf{x}')
\end{bmatrix} \\
\Sigma^{(l+1)}(\mathbf{x}, \mathbf{x}') &= \mathbb{E}_{f \sim \mathcal{N}(0, \lambda^{(l)})}[\sigma(f(\mathbf{x})) \sigma(f(\mathbf{x}'))] + \beta^2
\end{aligned}
\end{equation}
Here's a proof by mathematical induction:
(1) Let's begin with the case when $L=1$, which corresponds to a neural network with no nonlinearity function, and the input is processed by a simple affine transformation:
\begin{equation}
    \begin{aligned}
f(\mathbf{x};\theta) = \tilde{A}^{(1)}(\mathbf{x}) &= \frac{1}{\sqrt{n_0}}{\mathbf{w}^{(0)}}^\top\mathbf{x} + \beta\mathbf{b}^{(0)} \\
\text{where }\tilde{A}_m^{(1)}(\mathbf{x}) &= \frac{1}{\sqrt{n_0}}\sum_{i=1}^{n_0} w^{(0)}_{im}x_i + \beta b^{(0)}_m\quad \text{for }1 \leq m \leq n_1
\end{aligned}
\end{equation}
Since the weights and biases are initialized independently and identically distributed (i.i.d.), all output dimensions of this network, denoted as ${\tilde{A}^{(1)}1(\mathbf{x}), \dots, \tilde{A}^{(1)}{n_1}(\mathbf{x})}$, are also i.i.d. For different inputs, the $\textit{m}$-th network output $\tilde{A}^{(1)}_m(.)$ has a joint multivariate Gaussian distribution, which is equivalent to a Gaussian process with a covariance function. Here, we know that the mean $\mu_w=\mu_b=0$ and variance $\sigma^2_w = \sigma^2_b=1$.
\begin{equation}
    \begin{aligned}
\Sigma^{(1)}(\mathbf{x}, \mathbf{x}') 
&= \mathbb{E}[\tilde{A}_m^{(1)}(\mathbf{x})\tilde{A}_m^{(1)}(\mathbf{x}')] \\
&= \mathbb{E}\Big[\Big( \frac{1}{\sqrt{n_0}}\sum_{i=1}^{n_0} w^{(0)}_{i,m}x_i + \beta b^{(0)}_m \Big) \Big( \frac{1}{\sqrt{n_0}}\sum_{i=1}^{n_0} w^{(0)}_{i,m}x'_i + \beta b^{(0)}_m \Big)\Big] \\
&= \frac{1}{n_0} \sigma^2_w \sum_{i=1}^{n_0} \sum_{j=1}^{n_0} x_i{x'}_j + \frac{\beta \mu_b}{\sqrt{n_0}} \sum_{i=1}^{n_0} w_{im}(x_i + x'_i) + \sigma^2_b \beta^2 \\
&= \frac{1}{n_0}\mathbf{x}^\top{\mathbf{x}'} + \beta^2
\end{aligned}
\end{equation}
(2) Assuming that the proposition is true for $L=l$, a $l$-layer network, and thus $\tilde{A}^{(l)}_m(.)$ is a Gaussian process with covariance $\Sigma^{(l)}$ and ${\tilde{A}^{(l)}i}{i=1}^{n_l}$ are i.i.d., we can use induction to prove that the proposition is also true for $L=l+1$.
To do so, we compute the outputs as follows:
\begin{equation}
    \begin{aligned}
f(\mathbf{x};\theta) = \tilde{A}^{(l+1)}(\mathbf{x}) &= \frac{1}{\sqrt{n_l}}{\mathbf{w}^{(l)}}^\top \sigma(\tilde{A}^{(l)}(\mathbf{x})) + \beta\mathbf{b}^{(l)} \\
\text{where }\tilde{A}^{(l+1)}_m(\mathbf{x}) &= \frac{1}{\sqrt{n_l}}\sum_{i=1}^{n_l} w^{(l)}_{im}\sigma(\tilde{A}^{(l)}_i(\mathbf{x})) + \beta b^{(l)}_m \quad \text{for }1 \leq m \leq n_{l+1}
\end{aligned}
\end{equation}
We can conclude that the sum of contributions from the previous hidden layers has an expected value of zero:
\begin{equation}
    \begin{aligned}
\mathbb{E}[w^{(l)}_{im}\sigma(\tilde{A}^{(l)}_i(\mathbf{x}))] 
&= \mathbb{E}[w^{(l)}_{im}]\mathbb{E}[\sigma(\tilde{A}^{(l)}_i(\mathbf{x}))] 
= \mu_w \mathbb{E}[\sigma(\tilde{A}^{(l)}_i(\mathbf{x}))] = 0 \\
\mathbb{E}[\big(w^{(l)}_{im}\sigma(\tilde{A}^{(l)}_i(\mathbf{x}))\big)^2]
&= \mathbb{E}[{w^{(l)}_{im}}^2]\mathbb{E}[\sigma(\tilde{A}^{(l)}_i(\mathbf{x}))^2] 
= \sigma_w^2 \Sigma^{(l)}(\mathbf{x}, \mathbf{x})
= \Sigma^{(l)}(\mathbf{x}, \mathbf{x})
\end{aligned}
\end{equation}

Based on the assumption that ${\tilde{A}^{(l)}i(\mathbf{x})}{i=1}^{n_l}$ are i.i.d., we can apply the central limit theorem. As the hidden layer width increases infinitely, i.e., $n_l \to \infty$, we can conclude that $\tilde{A}^{(l+1)}_m(\mathbf{x})$ follows a Gaussian distribution with a variance of $\beta^2 + \text{Var}(\tilde{A}i^{(l)}(\mathbf{x}))$. It is worth noting that ${\tilde{A}^{(l+1)}1(\mathbf{x}), \dots, \tilde{A}^{(l+1)}{n{l+1}}(\mathbf{x})}$ remain i.i.d. Thus, $\tilde{A}^{(l+1)}_m(.)$ can be considered a Gaussian process with a covariance function given by:
\begin{equation}
    \begin{aligned}
\Sigma^{(l+1)}(\mathbf{x}, \mathbf{x}') 
&= \mathbb{E}[\tilde{A}^{(l+1)}_m(\mathbf{x})\tilde{A}^{(l+1)}_m(\mathbf{x}')] \\
&= \frac{1}{n_l} \sigma\big(\tilde{A}^{(l)}_i(\mathbf{x})\big)^\top \sigma\big(\tilde{A}^{(l)}_i(\mathbf{x}')\big) + \beta^2 \quad\text{;similar to how we get }\Sigma^{(1)}
\end{aligned}
\end{equation}

According to the central limit theorem, $\Sigma^{(l+1)}(\mathbf{x}, \mathbf{x}')  \to \mathbb{E}_{f \sim \mathcal{N}(0, \Lambda^{(l)})}[\sigma(f(\mathbf{x}))^\top \sigma(f(\mathbf{x}'))] + \beta^2$ when $n_l \to \infty$.
The Gaussian processes described in the previous section are known as Neural Network Gaussian Processes (NNGP).

\subsection{Remarks - Deterministic Neural Tangent Kernel}
\label{sec: key_results}

After the necessary preparations, we are now ready to examine the most important propositions from the NTK paper.
\begin{itemize}
    \item As the number of neurons in each layer $n_1, \dots, n_L$ approaches infinity, the neural tangent kernel, which is random at initialization, converges in
probability to a deterministic limit;
    \item As the number of neurons in each layer $n_1, \dots, n_L$ approaches infinity, stays constant during training.
\end{itemize}

In our paper, we extend them to zero-bias neural networks by simply examining the effect of the scalar constant $\beta = 0$ corresponding to bias terms, we then draw the following two remarks

\textbf{Remark 1}
\emph{
When the width of networks goes to infinite, both NTKs of zero-bias neural networks and normal neural networks converge in probability to the same deterministic limit (limiting kernel).}

\textbf{Remark 2}
\emph{
In the infinite-width limit, both NTKs of zero-bias neural networks and normal neural networks stay asymptotically the same constant during training.}

In this section, we provide the proof/analysis of Remark 1 only. For detailed proof/analysis of Remark 2, readers may refer to the original paper \cite{ntk2018}. The derivation of Remark 2 follows the same spirit as that of Remark 1, where we simply drop the bias terms $\mathbf{b}$ and its scalar constant $\beta$. \textbf{ This yields exactly the same form of equations. In fact, it is trivial to show that almost every property of the NTK of zero-bias neural networks and normal neural networks is identical by ignoring $\beta$ since it does not affect any other terms in almost every expression/equation.}

\textbf{Proof:}
The proof/analysis of Remark 1 depends on mathematical induction:

First, we observe that $K^{(0)}$ is always equal to zero.
When $L=1$, we can directly obtain the NTK representation. It is deterministic and does not depend on network initialization. As there is no hidden layer, there is no concept of infinite width to consider.
\begin{equation}
    \begin{aligned}
f(\mathbf{x};\theta) &= \tilde{A}^{(1)}(\mathbf{x}) = \frac{1}{\sqrt{n_0}} {\mathbf{w}^{(0)}}^\top\mathbf{x} + \beta\mathbf{b}^{(0)} \\
K^{(1)}(\mathbf{x}, \mathbf{x}';\theta) 
&= \Big(\frac{\partial f(\mathbf{x}';\theta)}{\partial \mathbf{w}^{(0)}}\Big)^\top \frac{\partial f(\mathbf{x};\theta)}{\partial \mathbf{w}^{(0)}} +
\Big(\frac{\partial f(\mathbf{x}';\theta)}{\partial \mathbf{b}^{(0)}}\Big)^\top \frac{\partial f(\mathbf{x};\theta)}{\partial \mathbf{b}^{(0)}} \\
&= \frac{1}{n_0} \mathbf{x}^\top{\mathbf{x}'} + \beta^2 = \Sigma^{(1)}(\mathbf{x}, \mathbf{x}')
\end{aligned}
\end{equation}
(2) Consider the case $L=1$,  we assume that a $l$-layer network with $\tilde{P}$ parameters in total, $\tilde{\theta} = (\mathbf{w}^{(0)}, \dots, \mathbf{w}^{(l-1)}, \mathbf{b}^{(0)}, \dots, \mathbf{b}^{(l-1)}) \in \mathbb{R}^{\tilde{P}}$, has a NTK converging to a deterministic limit when $n_1, \dots, n_{l-1} \to \infty$.
\begin{equation}
    K^{(l)}(\mathbf{x}, \mathbf{x}';\tilde{\theta}) = \nabla_{\tilde{\theta}} \tilde{A}^{(l)}(\mathbf{x})^\top \nabla_{\tilde{\theta}} \tilde{A}^{(l)}(\mathbf{x}') \to K^{(l)}_{\infty}(\mathbf{x}, \mathbf{x}')
\end{equation}
Note that $K_\infty^{(l)}$ is independent of $\theta$.
Now let's consider the case of $L=l+1$. Compared to an $l$-layer network, a $(l+1)$-layer network has an additional weight matrix $\mathbf{w}^{(l)}$ and bias vector $\mathbf{b}^{(l)}$, resulting in a total parameter vector of $\theta = (\tilde{\theta}, \mathbf{w}^{(l)}, \mathbf{b}^{(l)})$.
The output function of this $(l+1)$-layer network is:
\begin{equation}
    f(\mathbf{x};\theta) = \tilde{A}^{(l+1)}(\mathbf{x};\theta) = \frac{1}{\sqrt{n_l}} {\mathbf{w}^{(l)}}^\top \sigma\big(\tilde{A}^{(l)}(\mathbf{x})\big) + \beta \mathbf{b}^{(l)}
\end{equation}
We can easily compute its derivative with respect to different sets of parameters. Let us denote $\tilde{A}^{(l)}$ as $\tilde{A}^{(l)}(\mathbf{x})$ for brevity in the following equation:
\begin{equation}
    \begin{aligned}
\nabla_{{\mathbf{w}^{(l)}}} f(\mathbf{x};\theta) &= {
    \frac{1}{\sqrt{n_l}} \sigma\big(\tilde{A}^{(l)}\big)^\top
} \color{black}{\quad \in \mathbb{R}^{1 \times n_l}} \\
\nabla_{{\mathbf{b}^{(l)}}} f(\mathbf{x};\theta) &= { \beta } \\
\nabla_{\tilde{\theta}} f(\mathbf{x};\theta) &= \frac{1}{\sqrt{n_l}} \nabla_{\tilde{\theta}}\sigma(\tilde{A}^{(l)}) \mathbf{w}^{(l)} \\
&= {
    \frac{1}{\sqrt{n_l}}
    \begin{bmatrix}
        \dot{\sigma}(\tilde{A}_1^{(l)})\frac{\partial \tilde{A}_1^{(l)}}{\partial \tilde{\theta}_1} & \dots & \dot{\sigma}(\tilde{A}_{n_l}^{(l)})\frac{\partial \tilde{A}_{n_l}^{(l)}}{\partial \tilde{\theta}_1} \\
        \vdots \\       
        \dot{\sigma}(\tilde{A}_1^{(l)})\frac{\partial \tilde{A}_1^{(l)}}{\partial \tilde{\theta}_{\tilde{P}}}
        & \dots & \dot{\sigma}(\tilde{A}_{n_l}^{(l)})\frac{\partial \tilde{A}_{n_l}^{(l)}}{\partial \tilde{\theta}_{\tilde{P}}}\\
    \end{bmatrix}
    \mathbf{w}^{(l)}
    \color{black}{\quad \in \mathbb{R}^{\tilde{P} \times n_{l+1}}}
}
\end{aligned}
\end{equation}
where $\dot{\sigma}$ is the derivative of $\sigma$, and each entry in the matrix $\nabla_{\tilde{\theta}} f(\mathbf{x};\theta)$, located at $(p, m), 1 \leq p \leq \tilde{P}$ and $1 \leq m \leq n_{l+1}$, can be expressed as:
\begin{equation}
    \frac{\partial f_m(\mathbf{x};\theta)}{\partial \tilde{\theta}_p} = \sum_{i=1}^{n_l} w^{(l)}_{im} \dot{\sigma}\big(\tilde{A}_i^{(l)} \big) \nabla_{\tilde{\theta}_p} \tilde{A}_i^{(l)}
\end{equation}
The NTK for this ($l-1$)-layer network can be defined accordingly:
\begin{equation}
    \begin{aligned}
& K^{(l+1)}(\mathbf{x}, \mathbf{x}'; \theta) \\ 
=& \nabla_{\theta} f(\mathbf{x};\theta)^\top \nabla_{\theta} f(\mathbf{x};\theta) \\
=& \color{blue}{\nabla_{\mathbf{w}^{(l)}} f(\mathbf{x};\theta)^\top \nabla_{\mathbf{w}^{(l)}} f(\mathbf{x};\theta)} 
    + \color{green}{\nabla_{\mathbf{b}^{(l)}} f(\mathbf{x};\theta)^\top \nabla_{\mathbf{b}^{(l)}} f(\mathbf{x};\theta)}
    + \color{red}{\nabla_{\tilde{\theta}} f(\mathbf{x};\theta)^\top \nabla_{\tilde{\theta}} f(\mathbf{x};\theta)}  \\
=& \frac{1}{n_l} \Big[ 
    \color{blue}{\sigma(\tilde{A}^{(l)})\sigma(\tilde{A}^{(l)})^\top} 
    + \color{green}{\beta^2} \\
    &+
    \color{red}{
        {\mathbf{w}^{(l)}}^\top 
        \begin{bmatrix}
            \dot{\sigma}{(\tilde{A}_{1^{(l)}})}\dot{\sigma}{(\tilde{A}_{1^{(l)}})}\sum_{p=1}^{\tilde{P}} \frac{\partial {\tilde{A}_1^{(l)}}}{\partial {\tilde{\theta}_p}}\frac{\partial {\tilde{A}_1^{(l)}}}{\partial {\tilde{\theta}_p}} & \dots & \dot{\sigma}{(\tilde{A}_1^{(l)})}\dot{\sigma}{(\tilde{A}_{n_l}^{(l)})}\sum_{p=1}^{\tilde{P}} \frac{\partial {\tilde{A}_1^{(l)}}}{\partial {\tilde{\theta}_p}}\frac{\partial {\tilde{A}_{n_l}^{(l)}}}{\partial {\tilde{\theta}_p}} \\
            \vdots \\
            \dot{\sigma}(\tilde{A}_{n_l}^{(l)})\dot{\sigma}(\tilde{A}_1^{(l)})\sum_{p=1}^{\tilde{P}} \frac{\partial \tilde{A}_{n_l}^{(l)}}{\partial \tilde{\theta}_p}\frac{\partial \tilde{A}_1^{(l)}}{\partial \tilde{\theta}_p} & \dots & \dot{\sigma}(\tilde{A}_{n_l}^{(l)})\dot{\sigma}(\tilde{A}_{n_l}^{(l)})\sum_{p=1}^{\tilde{P}} \frac{\partial \tilde{A}_{n_l}^{(l)}}{\partial \tilde{\theta}_p}\frac{\partial \tilde{A}_{n_l}^{(l)}}{\partial \tilde{\theta}_p} \\
        \end{bmatrix}
        \mathbf{w}^{(l)}
    }
\color{black}{\Big]} \\
=& \frac{1}{n_l} \Big[ 
    \color{blue}{\sigma(\tilde{A}^{(l)})\sigma(\tilde{A}^{(l)})^\top} 
    + \color{green}{\beta^2} \\
    &+
    \color{red}{
        {\mathbf{w}^{(l)}}^\top 
        \begin{bmatrix}
            \dot{\sigma}(\tilde{A}_1^{(l)})\dot{\sigma}(\tilde{A}_1^{(l)})K^{(l)}_{11} & \dots & \dot{\sigma}(\tilde{A}_1^{(l)})\dot{\sigma}(\tilde{A}_{n_l}^{(l)})K^{(l)}_{1n_l} \\
            \vdots \\
            \dot{\sigma}(\tilde{A}_{n_l}^{(l)})\dot{\sigma}(\tilde{A}_1^{(l)})K^{(l)}_{n_l1} & \dots & \dot{\sigma}(\tilde{A}_{n_l}^{(l)})\dot{\sigma}(\tilde{A}_{n_l}^{(l)})K^{(l)}_{n_ln_l} \\
        \end{bmatrix}
        \mathbf{w}^{(l)}
    }
\color{black}{\Big]}
\end{aligned}
\end{equation}
where each individual entry at location $(m, n), 1 \leq m, n \leq n_{l+1}$, $(m, n), 1 \leq m, n \leq n_{l+1}$ of the matrix $K^{(l+1)}$ can be written as:
\begin{equation}
    \begin{aligned}
K^{(l+1)}_{mn} 
=& \frac{1}{n_l}\Big[
    \color{blue}{\sigma(\tilde{A}_m^{(l)})\sigma(\tilde{A}_n^{(l)})}
    + \color{green}{\beta^2} 
    + \color{red}{
    \sum_{i=1}^{n_l} \sum_{j=1}^{n_l} w^{(l)}_{im} w^{(l)}_{in} \dot{\sigma}(\tilde{A}_i^{(l)}) \dot{\sigma}(\tilde{A}_{j}^{(l)}) K_{ij}^{(l)}
}
\Big]
\end{aligned}
\end{equation}
When $n_l \to \infty$, the expression in blue and green approaches its limit (as shown in the previous section):
\begin{equation}
    \frac{1}{n_l}\sigma(\tilde{A}^{(l)})\sigma(\tilde{A}^{(l)}) + \beta^2\to \Sigma^{(l+1)}
\end{equation}
\textbf{Note that for zero-bias neural networks, we can simply drop $\beta^2$. In this case, they approach the same limit:}
\begin{equation}
    \frac{1}{n_l}\sigma(\tilde{A}^{(l)})\sigma(\tilde{A}^{(l)})\to \Sigma^{(l+1)}
\end{equation}
and the red section has the limit:
\begin{equation}
    \sum_{i=1}^{n_l} \sum_{j=1}^{n_l} w^{(l)}_{im} w^{(l)}_{in} \dot{\sigma}(\tilde{A}_i^{(l)}) \dot{\sigma}(\tilde{A}_{j}^{(l)}) K_{ij}^{(l)} 
\to
\sum_{i=1}^{n_l} \sum_{j=1}^{n_l} w^{(l)}_{im} w^{(l)}_{in} \dot{\sigma}(\tilde{A}_i^{(l)}) \dot{\sigma}(\tilde{A}_{j}^{(l)}) K_{\infty,ij}^{(l)}
\end{equation}
\textbf{ 
Note that for zero-bias neural networks, the red expression also converges to the same limit. Therefore, we conclude that both zero-bias neural networks and normal neural networks converge to the same deterministic limit (deterministic limiting kernel).}

\end{document}